\documentclass{article} % For LaTeX2e
\usepackage{preprint,times}

\usepackage{amsmath,amsfonts,bm}

\def\eqref#1{equation~\ref{#1}}
\def\1{\bm{1}}

\DeclareMathAlphabet{\mathsfit}{\encodingdefault}{\sfdefault}{m}{sl}
\SetMathAlphabet{\mathsfit}{bold}{\encodingdefault}{\sfdefault}{bx}{n}

\usepackage{hyperref}
\hypersetup{
  pdftitle={IneqAgent: Agentic Inequality Proving in Lean with Symbolic Computation},
  pdfauthor={Gaolei He, Jinhui Li, Ruobing Zuo, Xia Zeng, Zhengfeng Yang}
}
\usepackage{url}
\usepackage{amsthm}

\theoremstyle{remark}
\newtheorem{remark}{Remark}

\usepackage{graphicx}
\usepackage{booktabs}
\usepackage{multirow}
\usepackage{placeins}
\usepackage[skins]{tcolorbox}
\title{IneqAgent: Agentic Inequality Proving in Lean with Symbolic Computation}

\author{Gaolei He$^{1}$, Jinhui Li$^{1}$, Ruobing Zuo$^{1}$,
Xia Zeng$^{2}$, Zhengfeng Yang$^{1}$ \\
$^{1}$School of Software Engineering, East China Normal University \\
Shanghai, China \\
$^{2}$School of Computer and Information Science, Southwest University \\
Chongqing, China \\
{\small\texttt{\{glhe,jinhuili,rbzuo\}@stu.ecnu.edu.cn}} \\
{\small\texttt{xzeng0712@swu.edu.cn}, \texttt{zfyang@sei.ecnu.edu.cn}}
}

\begin{document}

\maketitle

\begin{abstract}
Formal inequality proving requires mathematical reasoning and exact
computation, often connected through intermediate lemmas and goal
reformulations. We introduce IneqAgent, an agentic framework combining
LLM reasoning with symbolic computation for inequality proving in
Lean~4. Lemma-level Monte Carlo Tree Search coordinates lemma
construction, goal reformulation, and symbolic backend selection.
Complementary sum-of-squares, Bernstein, and successive
difference-substitution backends produce exact certificates for
local Lean proof generation. The agent integrates these proofs with
LLM-generated steps and uses Lean feedback to construct complete
proofs checked against the original theorem statements.
On four benchmarks, IneqAgent with DeepSeek-V4-Flash solves
175 of 186 problems (94.09\%), achieving the highest overall success
rate among the evaluated methods.
Across four backbones, it achieves higher success rates at low
LLM inference cost thresholds than Direct Agent using the same
backbone, with lower mean inference costs over each method's
successful runs.
We also obtain Lean-verified strengthenings of Dittert's conjecture
in dimensions four and five, and certify the Holens--Djokovic
$(4,4)$ inequality, implying Merris's conjecture in dimension four. Code is available at \url{https://github.com/gaolei-he/IneqAgent}.

\end{abstract}

\section{Introduction}
\label{sec:introduction}

Inequality proving is an important task in automated mathematical
reasoning~\citep{wei2024aips,li2025lips}.
Even concise inequalities can require carefully chosen substitutions,
auxiliary bounds, and substantial algebraic computation.
Proving them in Lean~\citep{moura2021lean4} requires suitable
mathematical strategies and explicit justification of each step,
including sign and domain conditions.

Large language models (LLMs) support formal proof construction
through learned proof generation and interaction with tools and
verifiers~\citep{lin2025goedelv2,ren2025deepseekv2,
wang2025kiminaproverpreviewlargeformal,varambally2026hilbert,
requena2026minimal}.
Symbolic methods complement these capabilities by constructing
checkable algebraic certificates~\citep{peyrl2008rational,
kaltofen2012exact,magron2018realcertify}.
Combining them presents two challenges.
First, symbolic backends depend on the goal's algebraic form
and assumptions: a rational inequality, for example, may require
denominator positivity before polynomial reformulation.
Second, exact certificate construction can be computationally
demanding. When numerical solvers are used, floating-point
candidates typically satisfy certificate identities only
approximately and require exact recovery and verification.

We introduce IneqAgent, an agentic framework combining LLM reasoning,
symbolic computation, and Lean verification.
The agent uses lemma-level MCTS to explore
proof paths, choosing backend calls, goal reformulations, and
intermediate lemmas according to the goal's structure and backend
requirements.
It integrates certificate-derived local proofs with LLM-generated
steps, rewrites, and proofs of side conditions.
Lean checks the complete proof against the original theorem statement.

The symbolic module provides three complementary backends.
Using Putinar-type certificates~\citep{putinar1993positive},
the sum-of-squares (SOS) backend extends the numerical--symbolic
certification procedure of \citet{kaltofen2012exact} from
unconstrained problems to problems with equality and
inequality constraints through sequential recovery.
The Bernstein backend combines numerical and exact linear
programming with subdivision for box-constrained goals.
Successive difference substitution uses exact substitutions
and coefficient nonnegativity checks for polynomial goals over
nonnegative variables. The module supports applicable original goals and intermediate lemmas.

We apply IneqAgent to conjectures on matrix permanents, obtaining
Lean-verified proofs of strengthened inequalities for Dittert's
conjecture in dimensions four and five that establish quantitative
stability and unique maximizers.
We also verify an exact SOS certificate for the Holens--Djokovic
$(4,4)$ inequality in Lean and derive Merris's conjecture in
dimension four.
On four inequality benchmarks, IneqAgent with DeepSeek-V4-Flash
solves 175 of 186 problems (94.09\%), achieving the highest overall
success rate among the evaluated methods.
It also records lower mean solving times than the evaluated
LLM-based provers.
Across four backbones, it achieves lower mean LLM inference costs
and higher success rates at low cost thresholds than Direct Agent
using the same backbone.
Both means are computed over each method's own successful runs.
Ablations further support combining MCTS with symbolic computation.

Our main contributions are as follows:
\begin{itemize}
    \item
    We introduce IneqAgent, which uses lemma-level MCTS to
    coordinate LLM reasoning, goal reformulation, and symbolic
    computation, producing complete proofs checked by Lean
    against the original theorem statements.

    \item
    We develop a symbolic proving module with complementary
    SOS, Bernstein, and difference-substitution backends.
    We extend the recovery procedure of
    \citet{kaltofen2012exact} to constrained SOS certificates
    through sequential recovery, and translate certificates
    from all three backends into local Lean proofs.

    \item
   We demonstrate applications to matrix-permanent conjectures
   and introduce LitIneq, a benchmark of 74 formalized inequality
    tasks from research papers and formal-verification projects.
    We evaluate IneqAgent through benchmark comparisons,
    inference-cost analysis, and ablation studies.

\end{itemize}

\FloatBarrier

\section{Related Work}
\label{sec:related_work}

\paragraph{Symbolic Methods for Inequality Proving.}
General-purpose systems provide arithmetic and algebraic reasoning:
Z3 supports SMT-based arithmetic solving~\citep{demoura2008z3},
while Maple provides symbolic manipulation and quantifier
elimination~\citep{heck1993maple}.
Symbolic-numeric computation methods convert approximate
solutions into exact certificates~\citep{peyrl2008rational,
kaltofen2012exact,magron2018realcertify}.
Other approaches exploit domain structure, including Bernstein
representations with subdivision for box-constrained
problems~\citep{ray2009range} and successive difference
substitution for nonnegative variables~\citep{xia2016automated}.
Isabelle/HOL's \texttt{Sum\_of\_Squares} library integrates
SOS certificates with formal proof
checking~\citep{chaiebmeyer2025sos}.

\paragraph{LLMs for Formal Theorem Proving.}
 Neural provers generate tactics from formal proof states
and search over their continuations~\citep{polu2020generative,
lample2022hypertree}, with retrieval supplying relevant library
lemmas~\citep{yang2023leandojo}.
More recent models incorporate proof data synthesis,
self-correction, and subgoal decomposition~\citep{lin2025goedelv2,
ren2025deepseekv2}.
Agentic systems combine these capabilities with tool use and
verifier feedback.
Hilbert recursively combines informal reasoning with formal
subgoal proving~\citep{varambally2026hilbert}, while AxProverBase
uses a modular loop of proof proposals, review, memory, and
tool use~\citep{requena2026minimal}.

\paragraph{LLMs with Symbolic Reasoning for Inequality Proving.}
Existing systems assign different roles to learned reasoning
and symbolic computation.
LIPS combines symbolic generation and pruning of scaling steps
with LLM-generated rewrites and goal ranking, checking the
resulting proofs in Lean~\citep{li2025lips}.
IneqSearch combines symbolic decomposition with an LLM agent
that proposes transformations, and expands its theorem database
with newly proved inequalities~\citep{li2025ineqsearch}.
For unconstrained polynomial inequalities, NSPI uses an LLM
to conjecture approximate SOS decompositions, followed by
numerical refinement, rational recovery, and Lean proof
generation~\citep{zuo2026polyineq}.
IneqAgent uses lemma-level MCTS to coordinate intermediate
lemma construction, goal reformulation, and symbolic backend
selection according to the current assumptions and algebraic
structure. Its complementary backends produce exact
certificates that are translated into local Lean proofs and
combined with agent-generated steps to prove the original
theorem.

\FloatBarrier

\section{Methodology}
\label{headings}

\paragraph{Problem formulation.}
We study the problem of proving inequalities in real variables.
The expressions involved are constructed from real variables and
constants using arithmetic operations and elementary functions,
such as square roots, exponentials, logarithms, and trigonometric
functions.
For $\mathbf{x}=(x_1,\ldots,x_n) \in\mathbb{R}^{n}$, a typical
inequality-proving problem is to establish
\begin{equation}
f(\mathbf{x})\geq 0,
\qquad\forall \mathbf{x}\in D,
\label{eq:problem_formulation}
\end{equation}
where the domain $D$ is specified by the theorem to be proved, as follows:
\[
D:=\{\mathbf{x}\in\mathbb{R}^{n}\mid
\phi_i(\mathbf{x})\geq 0,\ i=1,\ldots,m;\
\psi_j(\mathbf{x})=0,\ j=1,\ldots,\ell\}.
\]
Here, $f$, $\phi_i$, and $\psi_j$ are real-valued expressions of the
form described above.
When $m=\ell=0$, we have $K=\mathbb{R}^{n}$, corresponding to an
unconstrained problem. Strict inequalities are also allowed in both the
constraints and the target.
The input is a theorem statement formalized in Lean~4,
and the task is to construct a complete proof accepted
by the Lean kernel.
For notational simplicity, we denote the Lean theorem-proving representation of problem (\ref{eq:problem_formulation}) as $\Gamma \vdash G$.

\begin{figure}[!h]
  \centering
  \includegraphics[width=\linewidth]{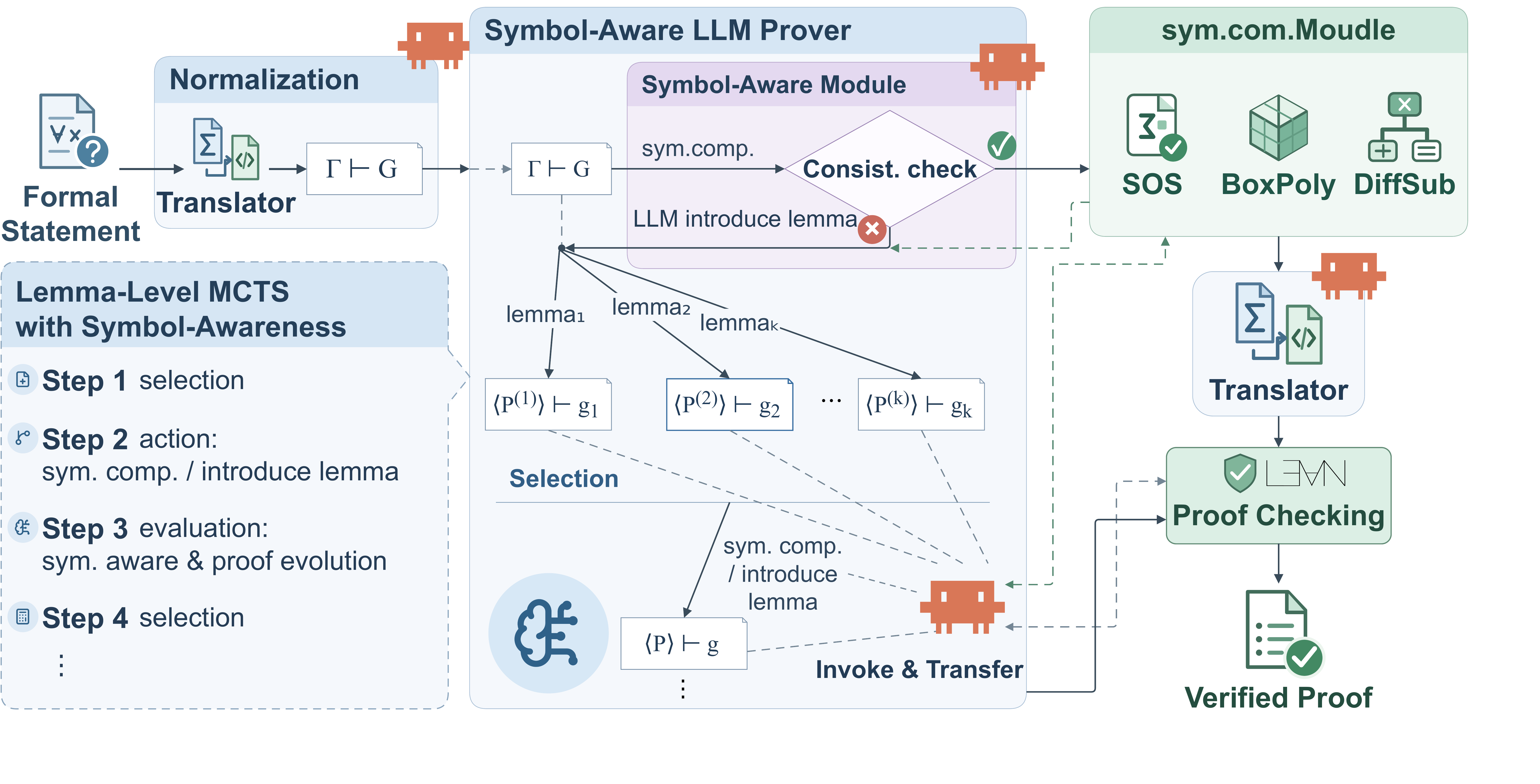}
  \caption{Overview of IneqAgent. The agent normalizes the formal statement
  and assesses each goal for symbolic computation. It develops intermediate
  lemmas when further reasoning is needed, coordinates tool calls, and
  integrates local proofs into the current proof context. Symbolic feedback
  and Lean diagnostics guide subsequent attempts until the original theorem
  is proved.}
  \label{fig:framework}
\end{figure}

To address problem (\ref{eq:problem_formulation}), we propose IneqAgent, an agentic framework for automated
inequality proving in Lean~4. The agent combines mathematical reasoning
with symbolic computation to reformulate goals, construct intermediate
lemmas, and assemble formal proofs. 
% Computational feedback and Lean
% diagnostics guide its subsequent proving steps. 
IneqAgent uses computational feedback from the symbolic computation module and diagnostics from the Lean proof module to guide and advance its subsequent proving steps.
The overall design is illustrated in Fig.~\ref{fig:framework}. We describe the framework in Section~\ref{sec:lemma_mcts} and the symbolic computation module in Section~\ref{sec:symbolic_proving}.

% \paragraph{Problem representation.}
% For real variables $\mathbf{x}=(x_1,\ldots,x_n)$, we represent an inequality
% problem as $\mathcal{P}=\langle f,\mathcal{C}\rangle$, with the target
% \begin{equation}
%   \forall\mathbf{x}\in\mathbb{R}^n,\qquad
%   \mathcal{C}(\mathbf{x})\Longrightarrow f(\mathbf{x})\geq0.
%   \label{eq:inequality_problem}
% \end{equation}
% Here $f$ is the difference between the two sides, and $\mathcal{C}$ contains
% the assumptions, including variable bounds, sign conditions, and equality
% constraints. The expression may contain polynomial, rational, or
% transcendental terms; strict inequalities use $f(\mathbf{x})>0$.
% During proving, a current goal is written as $\Gamma\vdash G$, where
% $\Gamma$ contains the variables, assumptions, and established intermediate
% facts, and $G$ is the goal to be proved. IneqAgent takes a Lean theorem
% statement as input and seeks a complete proof of that statement.

% Figure~\ref{fig:framework} shows how the agent normalizes the input,
% chooses proving actions, and uses tool feedback to extend the proof.
% Section~\ref{sec:lemma_mcts} describes this process, and
% Section~\ref{sec:symbolic_proving} presents the symbolic module.

\subsection{Agentic Framework for Inequality Proving}
\label{sec:lemma_mcts}

Assume that the inequality constraint problem to be proved is represented in Lean as $\Gamma \vdash G$. IneqAgent first normalizes the input problem into the standard form in (1). To facilitate more effective proof search, IneqAgent incorporates a dedicated symbolic computation module, \texttt{sym.com.Module} (see Section~3.2), while its core proof reasoning module is an LLM prover based on the \emph{lemma-level MCTS with symbolic awareness} framework. Based on the current proof state, IneqAgent determines whether to invoke \texttt{sym.com.Module}. When applicable, the module exploits algebraic computation to accelerate proof search and reduce the number of direct LLM reasoning calls, thereby improving computational efficiency. In addition, after each lemma introduction, reasoning step, or problem decomposition during the proof process, IneqAgent invokes Lean to formally verify the resulting proof state and ensure the correctness of the reasoning.

\paragraph{Lemma-Level MCTS with Symbol Awareness.}
The symbol-aware LLM prover adopts a lemma-level MCTS based proof search strategy, where each proof advancement can be viewed as an MCTS generation step consisting of \emph{selection}, \emph{action}, and \emph{evaluation}. Specifically, a proof state consists of a set of premises $\Gamma$, a target proposition $G$, and a context recording the current node state and its score, as detailed below. Starting from the initial proof state $\Gamma \vdash G$ as the root node, IneqAgent constructs an MCTS tree. It first selects the root node $v$ for expansion. The action begins with a symbolic-awareness check (described below). If the current form is not supported by the symbolic module, IneqAgent invokes the LLM to generate an appropriate lemma from the current proof state, and then invokes Lean to verify the lemma and derive a new proof state $\langle p^{(1)}\rangle \vdash g_1$ as a child node. Lean classifies the resulting state as \emph{open}, \emph{fail}, or \emph{solved}; if the node is verified as solved, the proof terminates. A state is \emph{solved} only when all proof obligations of the original theorem are discharged and Lean accepts the complete proof without \texttt{sorry} or \texttt{admit}; proving an intermediate lemma or resolving one subgoal leaves the state \emph{open} when other obligations remain. IneqAgent then evaluates the new child node using Lean checking results and LLM-based assessment of proof progress, and assigns a score to the node as described below. The scores of all nodes along the selected path are subsequently updated.

IneqAgent then performs another selection from the root node $v$ and continues with lemma introduction and node evaluation. To improve exploration of potentially promising proof paths, the root node is expanded whenever its number of accepted child nodes is below the current threshold $k$, according to a root-priority rule (see below). The current expansion threshold is given by $k:=B(v)$, where the design of $B(\cdot)$ is described in Appendix~A. If the proof has not yet been completed, a node $u$ is selected for the next expansion by descending through the highest-scoring open child at each level, as described below. IneqAgent performs symbolic awareness or lemma introduction on $u$ and evaluates the resulting node. This MCTS generation process is repeated until a node successfully passes Lean proof checking and is marked as solved. When a solved node is found, IneqAgent traverses the path from the root node to that node and extracts all intermediate proof steps to construct a complete Lean proof of the original problem.

\paragraph{Selection: Root-Priority Rule.}
During lemma-level MCTS based proof search, IneqAgent initializes a new root node and repeatedly expands it until it has $k:=B(v)$ child nodes. Each subsequent exploration begins at the root and selects the child with the highest score. If the selected node has fewer than $k$ children, it is expanded to generate an additional child node; otherwise, the selection operation identifies its highest-scoring child for the next exploration step.

\paragraph{Action: Symbol-Aware Actions.}
When an open node is selected for the first time, IneqAgent first performs symbolic awareness by checking whether the current theorem has an algebraic form supported by \texttt{sym.com.Module}. If the form is supported, IneqAgent directly invokes \texttt{sym.com.Module} to attempt an algebraic proof. If the form is unsupported or the symbolic computation fails, IneqAgent invokes the LLM to generate an appropriate lemma from the current proof state and adds it to the premise set. Lean is then invoked to verify the lemma and use it for subsequent reasoning. The resulting state can be \emph{fail} or \emph{solved}, or it can produce an open subgoal, which is recorded as an \emph{open} state. This process completes the expansion of a new node.

\paragraph{Evaluation: Lean Feedback and Proof-Evolution Scoring.}
The score of a newly expanded node $u$ follows the standard MCTS formulation (see Appendix~A). A key component is the mean reward
$
Q(u)=\frac{W(u)}{N(u)},
$
where $N(u)$ denotes the number of visits to node $u$, and $W(u)$ denotes its cumulative reward, updated as
$
W(u)\leftarrow W(u)+r(u).
$
The reward $r(u)$ is determined by the Lean-verified proof state of $u$:

$$
r(u)=
\begin{cases}
1, & \text{if $u$ is solved},\\
0, & \text{if $u$ fails},\\
\lambda r^{\mathrm{Lean}}_u+(1-\lambda)r^{\mathrm{LLM}}_u,
& \text{if $u$ is open}.
\end{cases}
$$

Here, $r^{\mathrm{Lean}}_u\in[0,1]$ is a heuristic progress score based on Lean checking results and proof-text features, including changes in unfinished markers and newly added structured proof steps, as detailed in Appendix~\ref{app:lemma_search}. In contrast, $r^{\mathrm{LLM}}_u\in[0,1]$ is obtained by comparing the parent and candidate proofs, using the candidate's Lean checking result to assess progress toward completing the original theorem. It rewards resolved proof obligations and useful intermediate results. IneqAgent then backpropagates the updated reward and updates the scores of all nodes along the selected path.

\subsection{Lean-Verified Symbolic Proving}
\label{sec:symbolic_proving}

The symbolic module addresses polynomial nonnegativity goals within the
general problem formulation above. These goals are prepared by the agent
and may coincide with the input theorem or arise from reformulations and
intermediate lemmas during proof search. Specifically, the module seeks
to establish
\begin{equation}
  p(\mathbf{x})\geq 0
  \qquad\text{for all }\mathbf{x}\in K,
  \label{eq:standard_polynomial_problem}
\end{equation}
where
\[
  K:=\{\mathbf{x}\in\mathbb{R}^{n}\mid
  g_i(\mathbf{x})\geq 0,\ i=1,\ldots,m;\
  h_j(\mathbf{x})=0,\ j=1,\ldots,\ell\},
\]
and $p,g_i,h_j$ are polynomials. The unconstrained case corresponds to
$m=\ell=0$. On success, the module returns an exact certificate for subsequent
Lean proof generation and verification. If no certificate is obtained,
it returns feedback to the agent for further proof search.

\paragraph{SOS-based proving under polynomial constraints.}
Motivated by Putinar's Positivstellensatz~\citep{putinar1993positive},
the SOS backend searches for a certificate of the form
\begin{equation}
  p=\sigma_0+\sum_{i=1}^{m}g_i\sigma_i
    +\sum_{j=1}^{\ell}h_jq_j,
  \label{eq:sos_certificate}
\end{equation}
where $\sigma_0,\ldots,\sigma_m$ are sums of squares and
$q_1,\ldots,q_\ell$ are polynomial multipliers unrestricted in sign.
Such an identity is sufficient to establish the nonnegativity of $p$
on $K$. Imposing degree bounds on the SOS polynomials and equality
multipliers yields a finite-dimensional SOS relaxation.
Each SOS polynomial is represented as
$\sigma_i=z_i^{\mathsf T}G_i z_i$, where $z_i$ is a monomial vector
and $G_i\succeq0$ is a Gram matrix.
Matching coefficients in the certificate identity gives linear
equality constraints on the Gram-matrix entries and the coefficients
of the equality multipliers. Together with the positive semidefiniteness
constraints, these define an SDP feasibility problem, which we solve
numerically using MOSEK.

For exact certification, we extend the hybrid symbolic-numeric recovery
procedure of \citet{kaltofen2012exact}, developed for unconstrained
problems, to the constrained SOS identity above.
Recovery proceeds sequentially: we first recover rational positive
semidefinite Gram matrices for the inequality SOS multipliers
$\sigma_1,\ldots,\sigma_m$ and keep these multipliers fixed.
The remaining task is to recover $\sigma_0$ and the equality
multipliers from
$p-\sum_{i=1}^{m}g_i\sigma_i
=\sigma_0+\sum_{j=1}^{\ell}h_jq_j$.
Gauss--Newton iterations refine the main SOS factorization and
the equality multipliers to reduce the residual in this remaining
identity. Rational recovery and exact coefficient correction then
seek a rational main Gram matrix $G_0$ and rational equality
multipliers satisfying the identity exactly.
The complete polynomial identity and the positive semidefiniteness
of all recovered Gram matrices are checked using exact arithmetic.
Only certificates passing these checks are returned for subsequent
Lean proof generation.
Details are provided in Appendix~\ref{app:sos_details}.

\paragraph{Bernstein-based proving under box constraints.}
For box-constrained goals, we use Bernstein representations and recursive
subdivision, following \citet{ray2009range}. For a box
$$K=\prod_{i=1}^{n}[a_i,b_i] \quad \text{with }\,\, a_i<b_i, $$ the affine substitution
$x_i=a_i+(b_i-a_i)y_i$ maps the problem to the unit box $[0,1]^n$.
We denote the transformed target polynomial by $\widetilde p$.
Choose a multidegree bound $\mathbf{d}=(d_1,\ldots,d_n)$ such that
$d_i\geq\deg_{y_i}\widetilde p$ for each $i$.
The Bernstein basis polynomials are
\begin{equation}
  B_{\mathbf{d},\alpha}(\mathbf{y})
  =\prod_{i=1}^{n}\binom{d_i}{\alpha_i}
    y_i^{\alpha_i}(1-y_i)^{d_i-\alpha_i},
  \qquad 0\leq\alpha_i\leq d_i.
  \label{eq:bernstein_basis}
\end{equation}
We seek a representation
$\widetilde p=\sum_{\alpha}c_\alpha B_{\mathbf{d},\alpha}$
with $c_\alpha\geq0$, where the sum ranges over the indices in
Eq.~\ref{eq:bernstein_basis}.
Let $A$ have the monomial coefficient vectors of these Bernstein
polynomials as its columns, and let $\mathbf{v}$ be the coefficient
vector of $\widetilde p$ in the same monomial basis.
Matching coefficients yields the LP feasibility problem
\begin{equation}
  \text{find }\mathbf{c}
  \quad\text{subject to}\quad
  A\mathbf{c}=\mathbf{v},\qquad \mathbf{c}\geq0.
  \label{eq:bernstein_lp}
\end{equation}
Any feasible solution certifies $\widetilde p\geq0$ on $[0,1]^n$,
since every Bernstein basis polynomial is nonnegative on this domain.
We first solve this feasibility problem in floating-point arithmetic
to obtain a numerical candidate.

If the numerical solve does not yield a suitable candidate certificate,
we select a variable and split its interval into two subintervals,
leaving the intervals of all other variables unchanged, and recursively
apply the procedure to both resulting subboxes. When a candidate is
obtained, we treat its near-zero entries as tentative zeros.
Fixing the corresponding coefficients to zero yields a reduced LP,
which we solve in exact arithmetic. If exact re-solving succeeds,
we check the complete coefficient-matching identity and the
nonnegativity of all recovered coefficients against the original
problem on the current box. A local certificate is accepted only
after these checks pass. The original box is certified once every
leaf box has an accepted certificate. The local certificates and the
subdivision tree are recorded for subsequent Lean proof generation,
including verification that the subboxes cover the original domain.

\paragraph{Successive difference substitution.}
For polynomial nonnegativity on the nonnegative orthant
$K=\mathbb{R}_{\geq0}^{n}$, DiffSub applies the successive difference
substitution method described by \citet[Chapter~10]{xia2016automated}.
For each permutation $\pi$ of $\{1,\ldots,n\}$, consider the ordering
$x_{\pi(1)}\geq\cdots\geq x_{\pi(n)}\geq0$ and introduce
nonnegative difference variables through
\begin{equation}
  x_{\pi(i)}=\sum_{j=i}^{n}t_j,
  \qquad i=1,\ldots,n,\qquad t_j\geq0.
  \label{eq:diffsub_substitution}
\end{equation}
Here, $t_i=x_{\pi(i)}-x_{\pi(i+1)}$ for $i<n$, and
$t_n=x_{\pi(n)}$. If the polynomial obtained by substituting into $p$
has only nonnegative coefficients, its nonnegativity follows from
$t_j\geq0$, certifying the current ordering branch.
Considering all variable orderings covers the entire domain $K$.

For every unresolved branch, the procedure repeats the ordering and
substitution on the new nonnegative variables; certified branches
require no further expansion. A complete certificate is obtained when
every leaf branch is certified. All substitutions and coefficient
checks use exact arithmetic, so no numerical recovery is required.
The certificate records the ordering cases, substitution identities,
and polynomial expansions at the leaves. 
If the search budget is exhausted while some branches remain
unresolved, the backend returns an unsolved result.

The Translator converts each exact certificate into a local Lean proof.
The agent integrates this proof into the surrounding context and
justifies any reformulation connecting the polynomial statement to
the original goal. A proof is accepted only after Lean verifies
the original theorem statement.

\section{Experiments}
\label{sec:experiments}

We evaluate IneqAgent on four inequality benchmarks, compare its
proving performance with existing systems, and examine LLM inference
cost across four backbones. We further present three case studies
on conjectures concerning matrix permanents. These case studies
are evaluated separately from the benchmark suite.

\subsection{Experimental Setup}
\label{sec:experimental_setup}

\paragraph{Benchmarks.}
The benchmark evaluation comprises 186 problems from three existing collections and our newly constructed LitIneq benchmark. MO-INT-20 is the 20-problem benchmark introduced with AIPS, comprising
algebraic inequalities from the IMO, IMO shortlists, and national
mathematical olympiads~\citep{wei2024aips}.
Ineq-Real is the 50-problem subset of Ineq-Comp drawn from mathematics
competitions, training materials, and Chinese Gaokao exams, with the problems
manually formalized in Lean~4~\citep{zhao2025ineqcomp}.
The Evan Chen benchmark consists of 42 problems selected from
\emph{A Brief Introduction to Olympiad Inequalities}, Evan Chen's 2014
notes~\citep{chen2014brief}.

To extend the evaluation beyond competition-oriented inequalities,
we construct LitIneq, a benchmark comprising 74 inequality-proving
tasks based on examples from research papers on inequality proving
and polynomial optimization, as well as selected inequality proof
obligations from the Flyspeck formalization of the Kepler conjecture.
For polynomial optimization examples, we formulate tasks that establish
bounds on the objective functions under the original constraints.
We formalize each task as a self-contained Lean~4 theorem statement.
Appendix~\ref{app:research_benchmark_sources} documents the source,
original theorem or example identifier, and formalization details
for each problem. Results are reported separately for each benchmark.
The conjecture case studies in
Section~\ref{sec:challenging_conjectures} are evaluated separately
and are excluded from the 186-problem aggregate.

\paragraph{Baselines and backbones.}
We compare with symbolic systems, formal provers, and proving agents.
The symbolic baselines are Z3~\citep{demoura2008z3}, Maple~\citep{heck1993maple}, and Isabelle/HOL's
\texttt{sos} method~\citep{chaiebmeyer2025sos}.
The formal provers and agents are
Goedel-Prover-V2-8B~\citep{lin2025goedelv2},
DeepSeek-Prover-V2-7B~\citep{ren2025deepseekv2},
AxProverBase~\citep{requena2026minimal}, LIPS~\citep{li2025lips},  and
Hilbert~\citep{varambally2026hilbert}.

IneqAgent uses DeepSeek-V4-Flash for the main benchmark comparison in Table~\ref{tab:main_results}.  
For the inference-cost comparison
in Table~\ref{tab:cost_comparison}, we evaluate IneqAgent and
Direct Agent with four backbones: DeepSeek-V4-Flash, Qwen3.8-Flash,
Kimi-K3, and GPT-5.5. Each pairwise comparison uses the same backbone. Direct Agent denotes Codex running with its default agent configuration and access to Lean.

\paragraph{Evaluation protocol.}
All benchmark methods are evaluated with a two-hour time limit
per problem. Success rates are computed over all problems in each
benchmark, counting unsolved and timed-out problems as failures.
Mean solving times are computed over the problems successfully
solved by each method. For Z3 and Maple, success is assessed from solver outputs, and 
these runs do not produce proofs checked by a proof assistant.
For IneqAgent, a problem is counted as solved only after Lean
accepts the generated proof. Detailed evaluation settings and
method-specific success criteria are provided in
Appendix~\ref{app:evaluation_protocol}.

\subsection{Main Results on Inequality Benchmarks}
\label{sec:main_results}

Table~\ref{tab:main_results} reports success rates and mean solving
times. IneqAgent achieves the highest success rate on each of the
four benchmarks, solving 19/20 problems in MO-INT-20, 46/50 in
Ineq-Real, 42/42 in Evan Chen, and 68/74 in LitIneq.
Overall, it solves 175/186 problems (94.09\%), compared with
141/186 (75.81\%) for Hilbert, the strongest baseline in overall
success rate, yielding an improvement of 18.28 percentage points.

On LitIneq, IneqAgent solves 68 problems, compared with 55 for
AxProverBase, the strongest baseline on this benchmark.
This corresponds to a success-rate improvement of 17.57 percentage 
points, extending the observed advantage beyond competition-oriented
inequalities to tasks drawn from research papers and Flyspeck.

\begin{table}[!htbp]
\caption{Success rate (SR, \%) and mean solving time over problems
successfully solved by each method (Time, seconds).
Overall aggregates all 186 problems; its mean time is weighted
by the number of successful problems in each benchmark.
Bold indicates the highest success rate in each benchmark
and overall. A dash indicates that no problem was solved.
}
  \label{tab:main_results}
  \centering
  \footnotesize
  \setlength{\tabcolsep}{2pt}
  \begin{tabular}{@{}l*{5}{rr}@{}}
    \toprule
    & \multicolumn{2}{c}{MO-INT-20}
    & \multicolumn{2}{c}{Ineq-Real}
    & \multicolumn{2}{c}{Evan Chen}
    & \multicolumn{2}{c}{LitIneq}
    & \multicolumn{2}{c}{Overall} \\
    \cmidrule(lr){2-3}\cmidrule(lr){4-5}\cmidrule(lr){6-7}
    \cmidrule(lr){8-9}\cmidrule(l){10-11}
    Method & SR & Time & SR & Time & SR & Time & SR & Time & SR & Time \\
    \midrule
    \multicolumn{11}{@{}l}{\textit{Symbolic computation}} \\[2pt]
    Z3 & 25.00 & 234.20 & 30.00 & 20.04 & 42.86 & 0.15 & 55.41 & 1.80 & 42.47 & 19.59 \\
    Maple & 0.00 & -- & 10.00 & 1.69 & 26.19 & 1.28 & 41.89 & 152.60 & 25.27 & 101.132 \\
    Isabelle \texttt{sos} & 0.00 & -- & 4.00 & 21.69 & 9.52 & 21.69 & 21.62 & 18.90 & 11.83 & 19.66 \\
    \midrule
    \multicolumn{11}{@{}l}{\textit{LLM provers}} \\[2pt]
    Goedel-Prover-V2-8B & 10.00 & 1323.86 & 28.00 & 779.01 & 30.95 & 842.92 & 59.46 & 856.65 & 39.25 & 852.11 \\
    DeepSeek-Prover-V2-7B & 10.00 & 503.94 & 18.00 & 492.84 & 28.57 & 422.14 & 44.59 & 525.31 & 30.11 & 497.22 \\
    \midrule
    \multicolumn{11}{@{}l}{\textit{LLM-based proving frameworks}} \\[2pt]
    AxProverBase & 75.00 & 1907.93 & 70.00 & 1381.69 & 73.81 & 1314.94 & 74.32 & 565.49 & 73.12 & 1094.43 \\
    Hilbert & 75.00 & 1945.00 & 86.00 & 2461.94 & 76.19 & 1848.77 & 68.92 & 875.19 & 75.81 & 1693.86 \\
    LIPS &35.00 &2479.74 &36.00 &2496.18 &54.76 &2165.82 &41.89 &109.207 &42.47 &1461.88 \\
    IneqAgent (ours) & \textbf{95.00} &333.87 & \textbf{92.00} &180.66 & \textbf{100.00} &147.92 & \textbf{91.89} &96.56 & \textbf{94.09} &156.76 \\
    \bottomrule
  \end{tabular}
\end{table}

\subsection{Proof Success under Cost Thresholds}
\label{sec:cost_comparison}
We examine whether IneqAgent enables more problems to be proved
at low LLM inference cost. Table~\ref{tab:cost_comparison} compares
IneqAgent with Direct Agent on all 186 problems using four backbones.
Each pairwise comparison uses the same backbone and pricing.

\paragraph{Cost accounting and thresholds.}
All costs are reported in US cents and measure LLM inference
expenditure only, excluding the computational costs of symbolic
solving and Lean verification.
The cost of a run accumulates across its LLM calls, including
unsuccessful attempts within that run.
For each threshold $b\in\{5,10,50,100\}$, we count the problems
successfully solved with cumulative LLM inference cost at most $b$.
Success rate uses all 186 problems as its denominator, while mean
cost is computed separately over each method's qualifying successful
runs. Unsolved and timed-out problems contribute to the success-rate
denominator but are excluded from the mean cost. The No cap setting includes all successful runs. These thresholds are applied post hoc rather than used as stopping
budgets  and every run remains subject to the two-hour time limit.

\begin{table}[!htbp]
\centering
\caption{Proof success and LLM inference cost on 186 problems.
Both methods use the backbone specified in the first column.
All thresholds and mean costs are in US cents.
SR (\%) is the percentage of all 186 problems solved with cumulative
LLM inference cost at or below the specified threshold; Cost is the
mean cost over those successful runs.
Thresholds are applied post hoc.
No cap includes all successful runs and the two-hour per-problem
limit still applies.
Within each backbone and threshold, the highest SR is bold;
the lower Cost is bold when both means are defined.}
\label{tab:cost_comparison}
\begingroup
\small
\setlength{\tabcolsep}{3pt}
\renewcommand{\arraystretch}{1.12}
\newcommand{\costTBD}{\textit{TBD}}
\resizebox{\linewidth}{!}{%
\begin{tabular}{ll*{5}{rr}}
\toprule
& & \multicolumn{10}{c}{LLM inference cost threshold (US cents)} \\
\cmidrule(lr){3-12}
& & \multicolumn{2}{c}{5}
  & \multicolumn{2}{c}{10}
  & \multicolumn{2}{c}{50}
  & \multicolumn{2}{c}{100}
  & \multicolumn{2}{c}{No cap} \\
\cmidrule(lr){3-4}\cmidrule(lr){5-6}\cmidrule(lr){7-8}
\cmidrule(lr){9-10}\cmidrule(lr){11-12}
Backbone & Method
& SR & Cost & SR & Cost & SR & Cost & SR & Cost & SR & Cost \\
\midrule
DeepSeek-V4-Flash & Direct Agent
& 66.67 & 1.9499
& 79.03 & 2.7609
& 90.32 & 4.6700
& 90.32 & 4.6700
& 90.32 & 4.6700 \\
& IneqAgent (ours)
& \textbf{88.17} & \textbf{0.6773}
& \textbf{92.47} & \textbf{0.9485}
& \textbf{94.09} & \textbf{1.3619}
& \textbf{94.09} & \textbf{1.3619}
& \textbf{94.09} & \textbf{1.3619} \\
\midrule
Qwen3.8-Flash & Direct Agent
& 68.28 & 1.7882
& 84.41 & 2.7370
& \textbf{92.47} & 4.0352
& \textbf{93.55} & 4.7061
& \textbf{93.55} & 4.7061 \\
& IneqAgent (ours)
& \textbf{83.87} & \textbf{0.8219}
& \textbf{89.25} & \textbf{1.1747}
& \textbf{92.47} & \textbf{1.5543}
& 92.47 & \textbf{1.5543}
& 92.47 & \textbf{1.5543} \\
\midrule
Kimi-K3 & Direct Agent
& 6.99 & 3.6009
& 25.27 & 6.3289
& 62.90 & 16.9118
& 76.34 & 25.8828
& \textbf{93.01} & 58.1366 \\
& IneqAgent (ours)
& \textbf{50.54} & \textbf{0.1310}
& \textbf{54.84} & \textbf{0.6915}
& \textbf{76.34} & \textbf{8.0921}
& \textbf{82.26} & \textbf{12.1085}
& 86.56 & \textbf{19.3747} \\
\midrule
GPT-5.5 & Direct Agent
& 0.00 & --
& 2.15 & 8.7023
& 57.53 & 31.1488
& \textbf{77.42} & 41.3875
& \textbf{90.86} & 72.3548 \\
& IneqAgent (ours)
& \textbf{48.92} & \textbf{0.1153}
& \textbf{50.54} & \textbf{0.3824}
& \textbf{67.74} & \textbf{7.0074}
& 75.27 & \textbf{13.2915}
& \textbf{90.86} & \textbf{63.7723} \\
\bottomrule
\end{tabular}%
}
\endgroup
\end{table}

IneqAgent achieves higher success rates at the 5- and 10-cent
thresholds across all four backbones. At 5 cents, the improvements
range from 15.59 to 48.92 percentage points. For instance,
DeepSeek-V4-Flash improves from 66.67\% to 88.17\%, while GPT-5.5
improves from 0.00\% to 48.92\%. At 50 cents, IneqAgent has
higher success rates for three backbones and ties Direct Agent with Qwen3.8-Flash.
These results show that more problems can be solved at low LLM
inference cost with IneqAgent.

The relative success rates vary at higher thresholds.
At 100 cents, Direct Agent achieves higher success rates with
Qwen3.8-Flash and GPT-5.5. Without a cost cap, IneqAgent achieves a higher success rate with DeepSeek-V4-Flash, matches Direct Agent with GPT-5.5, and has lower
success rates with Qwen3.8-Flash and Kimi-K3.
Mean inference costs over each method's successful runs are lower
for IneqAgent across all four backbones.
For Qwen3.8-Flash, the mean decreases from 4.7061 to 1.5543 cents, and 
for GPT-5.5, it decreases from 72.3548 to 63.7723 cents.
These means describe each method's own successful runs and do not
constitute comparisons on an identical set of problems.

\subsection{Ablation Study}
\label{sec:ablation_study}

We ablate lemma-level MCTS and symbolic computation using
DeepSeek-V4-Flash on the same 186 problems, with a two-hour
time limit per problem.
The \emph{Base agent} generates and revises proofs with Lean
feedback, using neither component.
\emph{Base + MCTS} adds lemma-level tree search, while
\emph{Base + Symbolic} enables symbolic proving of the original
goal and intermediate lemmas.
\emph{IneqAgent (full)} combines both components.

\begin{table}[!htbp]
  \caption{Ablation results with DeepSeek-V4-Flash on all 186
benchmark problems under a two-hour time limit per problem.
Mean LLM inference cost and wall-clock time are computed over
each variant's successful problems.
Bold indicates the highest success rate or the lowest reported mean.}
  \label{tab:ablations}
  \centering
  \small
  \begin{tabular}{@{}lrrr@{}}
    \toprule
    Variant & SR (\%) $\uparrow$ & Avg. LLM cost (US cents) $\downarrow$
      & Avg. time (s) $\downarrow$ \\
    \midrule
    Base agent & 90.32 & 4.6700 & 329.88 \\
    Base + MCTS & 91.40 & 2.2251 & 191.64 \\
    Base + Symbolic &91.40 & 2.8514 &200.56 \\
    IneqAgent (full) & \textbf{94.09} & \textbf{1.3619} & \textbf{156.76} \\
    \bottomrule
  \end{tabular}
\end{table}

Table~\ref{tab:ablations} shows that both components improve
performance when added individually. MCTS increases the number
of solved problems from 168 to 170, while symbolic computation
also increases it to 170. Combining both components solves 175 problems,
five more than either component alone. These results support the benefits of combining tree search
with symbolic computation in IneqAgent.

The full configuration also achieves the lowest mean LLM inference
cost and mean wall-clock time among the four variants.
Relative to the base agent, these means are lower by 70.84\%
and 52.48\%, respectively. Both measures are averaged over each
variant's own successful problems.

\subsection{Applications to Conjectures on Matrix Permanents}
\label{sec:challenging_conjectures}

We evaluate IneqAgent on three polynomial proof tasks arising
from conjectures on matrix permanents: strengthened forms of
Dittert's conjecture~\citep{minc1983theory} in dimensions four
and five, and the Holens--Djokovic $(4,4)$
inequality~\citep{holens1964doubly,djokovic1967waerden},
which implies the four-dimensional case of Merris's
conjecture~\citep{merris1973permanent}.
These case studies examine the framework's ability to combine
agentic reasoning with exact symbolic certification on
research-level mathematical problems.
They are evaluated separately from the 186-problem benchmark
suite and are not subject to its two-hour time limit.  Task sizes and costs are reported in
Table~\ref{tab:conjecture_complexity} in
Appendix~\ref{app:conjecture_proofs}.

\paragraph{Dittert's conjecture.}
For a nonnegative $n\times n$ matrix $A=(a_{ij})$ with
$\sum_{i,j}a_{ij}=n$, let $r_i=\sum_j a_{ij}$ and
$c_j=\sum_i a_{ij}$ denote its row and column sums.
Writing $\operatorname{per}(A)=
\sum_{\sigma\in S_n}\prod_{i=1}^{n}a_{i,\sigma(i)}$,
define $\Phi_n(A)=\prod_i r_i+\prod_j c_j-\operatorname{per}(A)$.
Dittert's conjecture asserts that
$\Phi_n(A)\leq2-n!/n^n$, with equality if and only if
$A=J_n$, where every entry of $J_n$ equals $1/n$.
For $n\in\{4,5\}$, IneqAgent produces Lean-verified proofs
of the strengthened polynomial inequalities
\begin{equation}
  p_n(A):=
  2-\frac{n!}{n^n}-\Phi_n(A)
  -\kappa_n\lVert A-J_n\rVert_F^2
  \geq0,
  \label{eq:dittert_stability}
\end{equation}
under the same constraints, where $\kappa_4=1/52$,
$\kappa_5=1/2500$, and $\lVert\cdot\rVert_F$ denotes
the Frobenius norm.
These bounds establish quantitative stability and the unique
maximizer $J_n$ in both dimensions.
For certificate search, we use the scaled target $P_4=32p_4$
in dimension four and a homogeneous reformulation
$\widetilde F_5$ in dimension five, defined in
Appendix~\ref{app:conjecture_proofs}.

\paragraph{Holens--Djokovic $(4,4)$ and Merris's conjecture.}
Let $\Omega_4$ denote the set of $4\times4$ nonnegative
matrices whose row and column sums equal one.
The Holens--Djokovic $(4,4)$ inequality is the polynomial
nonnegativity statement
\begin{equation}
  Q_4(A):=
  24\operatorname{per}(A)-2-\sum_{i,j=1}^{4}a_{ij}^{3}
  \geq0,
  \qquad A\in\Omega_4.
  \label{eq:hd44_polynomial}
\end{equation}
Expressing the remaining entries in terms of the upper-left
$3\times3$ block eliminates the row- and column-sum equalities,
yielding a degree-four polynomial problem in nine variables
while retaining the nonnegativity constraints on all entries.
IneqAgent constructs an exact rational SOS certificate
for the reduced problem. The argument for the contrary claim in
\citet[Theorem~6]{udayan2021merris}
contains an endpoint error(
see Remark~\ref{rem:hd44_prior_claim}).

This result implies Merris's conjecture in dimension four.
Let $S_j=\sum_{i=1}^{4}\operatorname{per}(A(i\mid j))$,
where $A(i\mid j)$ is obtained by deleting row $i$ and column $j$.
For $A\in\Omega_4$,
\[
  \min_{1\leq j\leq4}S_j
  \leq\frac14\sum_{j=1}^{4}S_j
  =4\operatorname{per}(A)-\frac16Q_4(A)
  \leq4\operatorname{per}(A).
\]
The SOS certificate and the case in which $S_4$ is minimal
are verified in Lean; column relabeling gives the conclusion
for any minimizing column.

% These case studies are evaluated separately from the benchmark's
% two-hour per-problem limit.
% Proof details and computational costs are provided in
% Appendix~\ref{app:conjecture_proofs}.

\section{Conclusion}
\label{sec:conclusion}

We introduced IneqAgent, an agentic framework for formal inequality
proving in Lean~4. Through lemma-level MCTS, it coordinates LLM
reasoning, goal reformulation, and symbolic computation, integrating
certificate-derived local proofs into complete proofs checked
against the original theorem statements.
Evaluation on four benchmarks and three polynomial proof tasks
arising from matrix-permanent conjectures demonstrates its
proving capability.
Across four backbones, IneqAgent achieves higher success rates
at low LLM inference cost thresholds than Direct Agent using
the same backbone, with lower mean inference costs over each
method's successful runs.

\subsection*{AI use statement}

We used generative AI tools to polish the writing, assist in finding relevant literature and reference materials, and carry out experiments according to author-specified instructions, including which tests to run and how to run them. We reviewed the AI-assisted text, consulted the cited sources, and checked the experimental outputs. We take full responsibility for the content and results of this work.

\section*{Acknowledgements}

This work was supported in part by the National Key Research and Development Program of China under Grant
2023YFA1009402, the Strategic Priority Research Program
of Chinese Academy of Sciences under Grant XDA0480501.

\clearpage
\bibliography{references}

\begin{thebibliography}{47}
\providecommand{\natexlab}[1]{#1}
\providecommand{\url}[1]{\texttt{#1}}
\expandafter\ifx\csname urlstyle\endcsname\relax
  \providecommand{\doi}[1]{doi: #1}\else
  \providecommand{\doi}{doi: \begingroup \urlstyle{rm}\Url}\fi

\bibitem[Chaieb \& Meyer(2025)Chaieb and Meyer]{chaiebmeyer2025sos}
Amine Chaieb and Philipp Meyer.
\newblock {Sum of Squares}: A decision procedure for universal multivariate
  real arithmetic using semidefinite programming.
\newblock Isabelle/{HOL} Library, theory \texttt{Sum\_of\_Squares}, 2025.
\newblock URL
  \url{https://isabelle.in.tum.de/website-Isabelle2025-2/dist/library/HOL/HOL-Library/Sum_of_Squares.html}.

\bibitem[Chen(2014)]{chen2014brief}
Evan Chen.
\newblock A brief introduction to olympiad inequalities.
\newblock \url{https://web.evanchen.cc/handouts/Ineq/en.pdf}, 2014.

\bibitem[Cheon \& Wanless(2005)Cheon and Wanless]{cheon2005survey}
Gi-Sang Cheon and Ian~M. Wanless.
\newblock An update on {Minc}'s survey of open problems involving permanents.
\newblock \emph{Linear Algebra and its Applications}, 403:\penalty0 314--342,
  2005.
\newblock \doi{10.1016/j.laa.2005.02.030}.

\bibitem[Cheon \& Wanless(2012)Cheon and Wanless]{cheon2012dittert}
Gi-Sang Cheon and Ian~M. Wanless.
\newblock Some results towards the {Dittert} conjecture on permanents.
\newblock \emph{Linear Algebra and its Applications}, 436\penalty0
  (4):\penalty0 791--801, 2012.
\newblock \doi{10.1016/j.laa.2010.08.041}.

\bibitem[de~Moura \& Bj{\o}rner(2008)de~Moura and Bj{\o}rner]{demoura2008z3}
Leonardo de~Moura and Nikolaj Bj{\o}rner.
\newblock {Z3}: An efficient {SMT} solver.
\newblock In \emph{Tools and Algorithms for the Construction and Analysis of
  Systems}, pp.\  337--340. Springer, 2008.

\bibitem[de~Moura \& Ullrich(2021)de~Moura and Ullrich]{moura2021lean4}
Leonardo de~Moura and Sebastian Ullrich.
\newblock The {Lean} 4 theorem prover and programming language.
\newblock In \emph{Automated Deduction---CADE 28}, pp.\  625--635. Springer,
  2021.

\bibitem[Djokovi{\'c}(1967)]{djokovic1967waerden}
D.~{\v Z}. Djokovi{\'c}.
\newblock On a conjecture by {van der Waerden}.
\newblock \emph{Matemati{\v c}ki Vesnik}, 4(19)\penalty0 (4):\penalty0
  272--276, 1967.

\bibitem[Emiris(1997)]{emiris1997general}
Ioannis~Z. Emiris.
\newblock A general solver based on sparse resultants: Numerical issues and
  kinematic applications.
\newblock Research Report 3110, INRIA, 1997.

\bibitem[Emiris(1994)]{emiris1994sparse}
Ioannis~Zacharias Emiris.
\newblock \emph{Sparse Elimination and Applications in kinematics}.
\newblock PhD thesis, University of California, Berkeley, 1994.

\bibitem[Faug\`ere(1999)]{faugere1999new}
Jean-Charles Faug\`ere.
\newblock A new efficient algorithm for computing {Gr\"obner} bases ({F4}).
\newblock \emph{Journal of Pure and Applied Algebra}, 139\penalty0
  (1--3):\penalty0 61--88, 1999.
\newblock \doi{10.1016/S0022-4049(99)00005-5}.
\newblock URL \url{https://doi.org/10.1016/S0022-4049(99)00005-5}.

\bibitem[Heck(1993)]{heck1993maple}
Andr{\'e} Heck.
\newblock \emph{Introduction to {MAPLE}}.
\newblock Springer, 1993.
\newblock \doi{10.1007/978-1-4684-0519-4}.

\bibitem[Holens(1964)]{holens1964doubly}
F.~Holens.
\newblock Two aspects of doubly stochastic matrices: Permutation matrices and
  the minimum of the permanent function.
\newblock Master's thesis, University of Manitoba, 1964.

\bibitem[Hwang(1986)]{hwang1986permanents}
Suk~Geun Hwang.
\newblock A note on a conjecture on permanents.
\newblock \emph{Linear Algebra and its Applications}, 76:\penalty0 31--44,
  1986.
\newblock \doi{10.1016/0024-3795(86)90212-0}.

\bibitem[Hwang(1987)]{hwang1987dittert}
Suk~Geun Hwang.
\newblock On a conjecture of {E. Dittert}.
\newblock \emph{Linear Algebra and its Applications}, 95:\penalty0 161--169,
  1987.
\newblock \doi{10.1016/0024-3795(87)90032-2}.

\bibitem[Iserles(1995)]{iserles1995personal}
Arieh Iserles.
\newblock Personal communication at the occasion of the {AMS--SIAM} summer
  seminar in applied mathematics.
\newblock Park City, Utah, 1995.

\bibitem[Kaltofen et~al.(2012)Kaltofen, Li, Yang, and Zhi]{kaltofen2012exact}
Erich~L. Kaltofen, Bin Li, Zhengfeng Yang, and Lihong Zhi.
\newblock Exact certification in global polynomial optimization via
  sums-of-squares of rational functions with rational coefficients.
\newblock \emph{Journal of Symbolic Computation}, 47\penalty0 (1):\penalty0
  1--15, 2012.

\bibitem[Lample et~al.(2022)Lample, Lacroix, Lachaux, Rodriguez, Hayat, Lavril,
  Ebner, and Martinet]{lample2022hypertree}
Guillaume Lample, Timoth{\'e}e Lacroix, Marie-Anne Lachaux, Aur{\'e}lien
  Rodriguez, Amaury Hayat, Thibaut Lavril, Gabriel Ebner, and Xavier Martinet.
\newblock Hypertree proof search for neural theorem proving.
\newblock \emph{Advances in Neural Information Processing Systems},
  35:\penalty0 26337--26349, 2022.

\bibitem[Li et~al.(2025{\natexlab{a}})Li, Li, Tang, Zhang, Yao, Si, Yang, Yang,
  and Ma]{li2025lips}
Zenan Li, Zhaoyu Li, Wen Tang, Xian Zhang, Yuan Yao, Xujie Si, Fan Yang, Kaiyu
  Yang, and Xiaoxing Ma.
\newblock Proving olympiad inequalities by synergizing {LLMs} and symbolic
  reasoning.
\newblock In \emph{International Conference on Learning Representations}, pp.\
  2304--2329, 2025{\natexlab{a}}.

\bibitem[Li et~al.(2025{\natexlab{b}})Li, Liao, and Ye]{li2025ineqsearch}
Zhaoqun Li, Beishui Liao, and Qiwei Ye.
\newblock {IneqSearch}: Hybrid reasoning for olympiad inequality proofs.
\newblock In \emph{Advances in Neural Information Processing Systems},
  volume~38, 2025{\natexlab{b}}.
\newblock URL
  \url{https://proceedings.neurips.cc/paper_files/paper/2025/hash/6c92839f0f9cddc96c694712a7143b09-Abstract-Conference.html}.

\bibitem[Lin et~al.(2025)Lin, Tang, Lyu, Yang, Chung, Zhao, Jiang, Geng, Ge,
  Sun, et~al.]{lin2025goedelv2}
Yong Lin, Shange Tang, Bohan Lyu, Ziran Yang, Jui-Hui Chung, Haoyu Zhao, Lai
  Jiang, Yihan Geng, Jiawei Ge, Jingruo Sun, et~al.
\newblock Goedel-prover-v2: Scaling formal theorem proving with scaffolded data
  synthesis and self-correction.
\newblock \emph{arXiv preprint arXiv:2508.03613}, 2025.

\bibitem[Liu et~al.(2025)Liu, Qi, Zeng, Dehbi, and Yang]{liu2025appirl}
Banglong Liu, Niuniu Qi, Xia Zeng, Lydia Dehbi, and Zhengfeng Yang.
\newblock Automated proof of polynomial inequalities via reinforcement
  learning.
\newblock \emph{arXiv preprint arXiv:2503.06592}, 2025.
\newblock URL \url{https://arxiv.org/abs/2503.06592}.

\bibitem[Magron \& Safey El~Din(2018)Magron and Safey
  El~Din]{magron2018realcertify}
Victor Magron and Mohab Safey El~Din.
\newblock {RealCertify}: A {Maple} package for certifying non-negativity.
\newblock \emph{ACM Communications in Computer Algebra}, 52\penalty0
  (2):\penalty0 34--37, 2018.

\bibitem[Marcus \& Merris(1973)Marcus and Merris]{marcus1973adjoints}
Marvin Marcus and Russell Merris.
\newblock A relation between the permanental and determinantal adjoints.
\newblock \emph{Journal of the Australian Mathematical Society}, 15\penalty0
  (3):\penalty0 270--271, 1973.
\newblock \doi{10.1017/S1446788700013173}.

\bibitem[Merris(1973)]{merris1973permanent}
Russell Merris.
\newblock The permanent of a doubly stochastic matrix.
\newblock \emph{The American Mathematical Monthly}, 80\penalty0 (7):\penalty0
  791--793, 1973.
\newblock \doi{10.1080/00029890.1973.11993372}.

\bibitem[Minc(1983)]{minc1983theory}
Henryk Minc.
\newblock Theory of permanents 1978--1981.
\newblock \emph{Linear and Multilinear Algebra}, 12:\penalty0 227--263, 1983.

\bibitem[Mu{\~n}oz \& Narkawicz(2013)Mu{\~n}oz and
  Narkawicz]{munoz2013bernstein}
C{\'e}sar Mu{\~n}oz and Anthony Narkawicz.
\newblock Formalization of {Bernstein} polynomials and applications to global
  optimization.
\newblock \emph{Journal of Automated Reasoning}, 51\penalty0 (2):\penalty0
  151--196, 2013.
\newblock \doi{10.1007/s10817-012-9256-3}.
\newblock URL \url{https://doi.org/10.1007/s10817-012-9256-3}.

\bibitem[Noonburg(1989)]{noonburg1989neural}
V.~W. Noonburg.
\newblock A neural network modeled by an adaptive {Lotka--Volterra} system.
\newblock \emph{SIAM Journal on Applied Mathematics}, 49\penalty0 (6):\penalty0
  1779--1792, 1989.
\newblock \doi{10.1137/0149109}.
\newblock URL \url{https://doi.org/10.1137/0149109}.

\bibitem[Pang(2026)]{pang2026dittert}
Zhekai Pang.
\newblock Proof of {Dittert}'s conjecture for dimensions {$n\geq17$}.
\newblock \emph{arXiv preprint arXiv:2606.01531}, 2026.
\newblock URL \url{https://arxiv.org/abs/2606.01531}.

\bibitem[Peyrl \& Parrilo(2008)Peyrl and Parrilo]{peyrl2008rational}
Helfried Peyrl and Pablo~A. Parrilo.
\newblock Computing sum of squares decompositions with rational coefficients.
\newblock \emph{Theoretical Computer Science}, 409\penalty0 (2):\penalty0
  269--281, 2008.

\bibitem[Polu \& Sutskever(2020)Polu and Sutskever]{polu2020generative}
Stanislas Polu and Ilya Sutskever.
\newblock Generative language modeling for automated theorem proving.
\newblock \emph{arXiv preprint arXiv:2009.03393}, 2020.

\bibitem[Putinar(1993)]{putinar1993positive}
M.~Putinar.
\newblock Positive polynomials on compact semi-algebraic sets.
\newblock \emph{Indiana University Mathematics Journal}, 42\penalty0
  (3):\penalty0 969--984, 1993.

\bibitem[Ray \& Nataraj(2009)Ray and Nataraj]{ray2009range}
Shashwati Ray and P.~S.~V. Nataraj.
\newblock An efficient algorithm for range computation of polynomials using the
  {Bernstein} form.
\newblock \emph{Journal of Global Optimization}, 45\penalty0 (3):\penalty0
  403--426, 2009.
\newblock \doi{10.1007/s10898-008-9382-y}.
\newblock URL \url{https://doi.org/10.1007/s10898-008-9382-y}.

\bibitem[Ren et~al.(2025)Ren, Shao, Song, Xin, Wang, Zhao, Zhang, Fu, Zhu,
  Yang, et~al.]{ren2025deepseekv2}
Z.~Z. Ren, Zhihong Shao, Junxiao Song, Huajian Xin, Haocheng Wang, Wanjia Zhao,
  Liyue Zhang, Zhe Fu, Qihao Zhu, Dejian Yang, et~al.
\newblock Deepseek-prover-v2: Advancing formal mathematical reasoning via
  reinforcement learning for subgoal decomposition.
\newblock \emph{arXiv preprint arXiv:2504.21801}, 2025.

\bibitem[Requena et~al.(2026)Requena, Letson, Nowakowski, Beltran-Ferreiro, and
  Sarra]{requena2026minimal}
Borja Requena, Austin Letson, Krystian Nowakowski, Izan Beltran-Ferreiro, and
  Leopoldo Sarra.
\newblock A minimal agent for automated theorem proving.
\newblock \emph{arXiv preprint arXiv:2602.24273}, 2026.

\bibitem[Sinkhorn(1984)]{sinkhorn1984dittert}
Richard Sinkhorn.
\newblock A problem related to the van der {Waerden} permanent theorem.
\newblock \emph{Linear and Multilinear Algebra}, 16\penalty0 (1--4):\penalty0
  167--173, 1984.
\newblock \doi{10.1080/03081088408817620}.

\bibitem[Udayan \& Somasundaram(2021)Udayan and Somasundaram]{udayan2021merris}
Divya~K. Udayan and Kanagasabapathi Somasundaram.
\newblock The inequalities of {Merris} and {Foregger} for permanents.
\newblock \emph{Symmetry}, 13\penalty0 (10):\penalty0 1782, 2021.
\newblock \doi{10.3390/sym13101782}.

\bibitem[Udayan \& Somasundaram(2024)Udayan and Somasundaram]{udayan2024lih}
Divya~K. Udayan and Kanagasabapathi Somasundaram.
\newblock {Lih Wang}'s and {Dittert}'s conjectures on permanents.
\newblock \emph{Special Matrices}, 12\penalty0 (1):\penalty0 20240006, 2024.
\newblock \doi{10.1515/spma-2024-0006}.

\bibitem[Varambally et~al.(2026)Varambally, Voice, Sun, Chen, Yu, and
  Ye]{varambally2026hilbert}
Sumanth Varambally, Thomas Voice, Yanchao Sun, Zhifeng Chen, Rose Yu, and
  Ke~Ye.
\newblock Hilbert: Recursively building formal proofs with informal reasoning.
\newblock In \emph{The Fourteenth International Conference on Learning
  Representations}, 2026.

\bibitem[Verschelde(1996)]{verschelde1996phc}
Jan Verschelde.
\newblock \emph{Polynomial homotopy continuation: A portable {Ada} software
  package}, 1996.
\newblock URL \url{ftp://ftp.cs.kuleuven.ac.be/pub/NumAnal-ApplMath/PHC}.
\newblock Computer program documentation.

\bibitem[Verschelde(2001)]{verschelde2001phc}
Jan Verschelde.
\newblock The {PHC} pack, the database of polynomial systems.
\newblock Technical report, University of Illinois, Mathematics Department,
  Chicago, USA, 2001.

\bibitem[Wang et~al.(2025)Wang, Unsal, Lin, Baksys, Liu, Santos, Sung, Vinyes,
  Ying, Zhu, Lu, de~Saxcé, Bailey, Song, Xiao, Zhang, Zhang, Pu, Zhu, Liu,
  Bayer, Michel, Yu, Dreyfus-Schmidt, Tunstall, Pagani, Machado, Bourigault,
  Wang, Polu, Barroyer, Li, Niu, Fleureau, Hu, Yu, Wang, Yang, Liu, and
  Li]{wang2025kiminaproverpreviewlargeformal}
Haiming Wang, Mert Unsal, Xiaohan Lin, Mantas Baksys, Junqi Liu, Marco~Dos
  Santos, Flood Sung, Marina Vinyes, Zhenzhe Ying, Zekai Zhu, Jianqiao Lu,
  Hugues de~Saxcé, Bolton Bailey, Chendong Song, Chenjun Xiao, Dehao Zhang,
  Ebony Zhang, Frederick Pu, Han Zhu, Jiawei Liu, Jonas Bayer, Julien Michel,
  Longhui Yu, Léo Dreyfus-Schmidt, Lewis Tunstall, Luigi Pagani, Moreira
  Machado, Pauline Bourigault, Ran Wang, Stanislas Polu, Thibaut Barroyer,
  Wen-Ding Li, Yazhe Niu, Yann Fleureau, Yangyang Hu, Zhouliang Yu, Zihan Wang,
  Zhilin Yang, Zhengying Liu, and Jia Li.
\newblock Kimina-prover preview: Towards large formal reasoning models with
  reinforcement learning, 2025.
\newblock URL \url{https://arxiv.org/abs/2504.11354}.

\bibitem[Wei et~al.(2024)Wei, Sun, and Wang]{wei2024aips}
Chenrui Wei, Mengzhou Sun, and Wei Wang.
\newblock Proving olympiad algebraic inequalities without human demonstrations.
\newblock \emph{Advances in Neural Information Processing Systems}, 37, 2024.

\bibitem[Wright(1985)]{wright1985finding}
Alden~H. Wright.
\newblock Finding all solutions to a system of polynomial equations.
\newblock \emph{Mathematics of Computation}, 44\penalty0 (169):\penalty0
  125--133, 1985.
\newblock \doi{10.1090/S0025-5718-1985-0771035-4}.
\newblock URL \url{https://doi.org/10.1090/S0025-5718-1985-0771035-4}.

\bibitem[Xia \& Yang(2016)Xia and Yang]{xia2016automated}
Bican Xia and Lu~Yang.
\newblock \emph{Automated Inequality Proving and Discovering}.
\newblock World Scientific, 2016.

\bibitem[Yang et~al.(2023)Yang, Swope, Gu, Chalamala, Song, Yu, Godil, Prenger,
  and Anandkumar]{yang2023leandojo}
Kaiyu Yang, Aidan~M. Swope, Alex Gu, Rahul Chalamala, Peiyang Song, Shixing Yu,
  Saad Godil, Ryan Prenger, and Anima Anandkumar.
\newblock {LeanDojo}: Theorem proving with retrieval-augmented language models.
\newblock In \emph{Advances in Neural Information Processing Systems}, 2023.

\bibitem[Zhao et~al.(2025)Zhao, Geng, Tang, Lin, Lyu, Lin, Jin, and
  Arora]{zhao2025ineqcomp}
Haoyu Zhao, Yihan Geng, Shange Tang, Yong Lin, Bohan Lyu, Hongzhou Lin, Chi
  Jin, and Sanjeev Arora.
\newblock {Ineq-Comp}: Benchmarking human-intuitive compositional reasoning in
  automated theorem proving on inequalities.
\newblock In \emph{Advances in Neural Information Processing Systems}, 2025.
\newblock URL \url{https://arxiv.org/abs/2505.12680}.
\newblock Datasets and Benchmarks Track.

\bibitem[Zuo et~al.(2026)Zuo, Zhao, He, Yang, and Wang]{zuo2026polyineq}
Ruobing Zuo, Hanrui Zhao, Gaolei He, Zhengfeng Yang, and Jianlin Wang.
\newblock From {LLM}-generated conjectures to {Lean} formalizations: Automated
  polynomial inequality proving via sum-of-squares certificates.
\newblock OpenReview: \url{https://openreview.net/forum?id=Lz8rHDmmKt}, 2026.
\newblock URL \url{https://arxiv.org/abs/2605.15445}.
\newblock Accepted to ICML 2026.

\end{thebibliography}
\bibliographystyle{references}

\clearpage
\appendix

\section{Lemma-Level Search Details}
\label{app:lemma_search}

This appendix gives the search rules used to allocate proving attempts
in Section~\ref{sec:lemma_mcts}. The root node is denoted by $v$. For a
selected node, we use $u$, and $u'$ denotes a newly generated child or
attempted child. Each node $u$ represents a partial proof state
$\Gamma_u\vdash G_u$, comprising the local context, established facts,
and remaining goals. The initial root state is
$\Gamma_v\vdash G_v$, corresponding to the initial state
$\Gamma\vdash G$. An accepted edge
$u\rightarrow u'$ records a proof extension with one or more fully
proved local lemmas. Paths represent successive extensions, while
sibling nodes retain alternative ways to advance the same state.

\paragraph{Selection and progressive widening.}
MCTS allocates proving attempts across alternative proof paths. For each
node $u$, we maintain a visit count $N(u)$ and accumulated reward $W(u)$
from attempts backed up through that node. Their ratio
$Q(u)=W(u)/N(u)$ is the mean reward, initialized to zero for unvisited
nodes. A stagnation count $S(u)$ tracks consecutive attempts to expand $u$
without useful progress. An open node is a valid proof state with remaining
goals. At each iteration, selection starts at the root and either expands
the current node or descends to an open child. Progressive widening sets
the threshold for adding children:
\begin{equation}
  B(u)=\max\!\left\{1,
  \left\lfloor \kappa\,\max(1,N(u))^{\alpha}\right\rfloor
  +\delta(S(u))\right\},
  \label{eq:progressive_widening}
\end{equation}
where $\kappa$ sets the branching scale, $\alpha\in(0,1)$ controls its growth,
and $\delta(S)=\min\{2,\lfloor S/2\rfloor\}$ allows up to two additional children
when progress stalls. Only accepted proof states count toward this child
limit; failed attempts are recorded separately. If $u$ has fewer than $B(u)$ children, the
agent generates a new extension from $u$. Otherwise, it descends to the open
child $u'$ maximizing the selection score
\begin{equation}
  U(u'\mid u)=Q(u')+c\sqrt{\frac{\log(N(u)+1)}{N(u')+1}}
  -\min\{\rho S(u'),\rho_{\max}\}.
  \label{eq:selection_score}
\end{equation}
Here $Q(u')$ favors branches with higher observed rewards, the exploration
term weighted by $c$ encourages less-visited branches, and the bounded
penalty discourages repeatedly unproductive ones, with strength $\rho$ and
maximum value $\rho_{\max}$. Selection continues until
a node can be expanded. If no open child is available, the agent attempts
a new extension at the current node $u$, even if the widening threshold
has been reached.

\paragraph{Search parameters.}
Table~\ref{tab:search_parameters} gives the default search configuration.
We use $\rho$ for the stagnation penalty and reserve $\lambda$ for the
Lean--LLM reward weight. New nodes initialize $N=W=Q=S=0$.
Selection scores are rounded to three decimal places.

\begin{table}[!htbp]
  \centering
  \caption{Default parameters for lemma-level search.}
  \label{tab:search_parameters}
  \begin{tabular}{@{}llr@{}}
    \toprule
    Parameter & Meaning & Value \\
    \midrule
    $\kappa$ & Progressive-widening scale & $1.0$ \\
    $\alpha$ & Progressive-widening exponent & $0.5$ \\
    $c$ & Exploration coefficient & $1.414$ \\
    $\lambda$ & Weight of the Lean-based reward & $0.5$ \\
    $\rho$ & Stagnation penalty per attempt & $0.06$ \\
    $\rho_{\max}$ & Maximum stagnation penalty & $0.30$ \\
    \bottomrule
  \end{tabular}
\end{table}

\paragraph{Reward and backpropagation.}
After expansion, Lean verification determines the candidate's status:
\texttt{solved} if it
completes the theorem without \texttt{sorry} or \texttt{admit},
\texttt{open} if its intermediate lemmas are verified but goals remain, or
\texttt{failed} if verification fails. Failed attempts provide diagnostics
for repair; only valid open states can be selected for further search.
For an open candidate $u'$, we combine a Lean-based heuristic score
$r^{\mathrm{Lean}}_{u'}\in[0,1]$ with an LLM score
$r^{\mathrm{LLM}}_{u'}\in[0,1]$. The former uses Lean checking results and
proof-text features defined below. The latter
compares the parent and candidate goals to estimate whether the new state
is simpler and closer to completing the theorem. The LLM score guides
search; correctness is determined by Lean. The reward is
\begin{equation}
  r(u')=
  \begin{cases}
    1, & \text{if }u'\text{ is solved},\\
    0, & \text{if }u'\text{ fails Lean verification},\\
    \lambda r^{\mathrm{Lean}}_{u'}+(1-\lambda)r^{\mathrm{LLM}}_{u'},
      & \text{if }u'\text{ is open},
  \end{cases}
  \label{eq:combined_reward}
\end{equation}
where $\lambda=0.5$ gives equal weight to the two progress signals for an
open candidate. The terminal rules take precedence: solved candidates
receive one and failed candidates receive zero, regardless of the
heuristic components. This final reward, rather than either component
alone, updates the search statistics.

A valid extension creates a child, and its reward is backed up from that
child to the root. A failed extension creates no accepted child; its
diagnostics are retained, and its zero reward is backed up from the
expanded parent $u$ to the root. Thus, failed attempts increase visit
counts without adding reward or occupying a child slot. For a failed
extension, $u'$ denotes the attempted child and $r(u')=0$. Let $u_i$
denote the $i$th node on the corresponding backpropagation path. Each
node $u_i$ is updated once:
\begin{equation}
  N(u_i)\leftarrow N(u_i)+1,\qquad
  W(u_i)\leftarrow W(u_i)+r(u'),\qquad
  Q(u_i)\leftarrow W(u_i)/N(u_i).
  \label{eq:backup}
\end{equation}
Thus, $Q(u_i)$ summarizes rewards from attempts through the branch rooted at
$u_i$, rather than assigning a fixed score to its current goals. Search
returns a complete Lean-verified proof when one is found and otherwise
terminates when the budget is exhausted or no open state remains.

\paragraph{Lean-based heuristic score.}
Each candidate undergoes a compilation check allowing unfinished
\texttt{sorry}/\texttt{admit} markers and a completion check requiring
a Lean-accepted proof without these markers. For the heuristic, let
$e$ indicate whether the proof text differs from its parent's, let
$m_u,m_{u'}$ count occurrences of the unfinished markers in the two
texts, and let $\Delta=m_u-m_{u'}$. Let $b$ indicate successful compilation.
These marker counts are textual features, rather than counts of Lean's
remaining goals.

Let $a$ count newly added lines beginning with \texttt{have},
\texttt{suffices}, a typed \texttt{let}, \texttt{lemma}, or
\texttt{theorem}. Added lines are candidate lines absent from the parent
text. These features measure proof structure; local proof validity is
determined separately by Lean. Starting from $h=0$, we add the applicable
terms in Table~\ref{tab:lean_reward_terms} and set
\begin{equation}
 r^{\mathrm{Lean}}_{u'}=
 \operatorname{round}_{3}\!\left(\min\{0.95,\max\{0,h\}\}\right).
 \label{eq:lean_heuristic_reward}
\end{equation}
The structured-step bonus and the compilable-edit bonus are mutually
exclusive. The agent timeout and execution-error penalties are also
mutually exclusive; other applicable terms accumulate. The heuristic
table includes failed-candidate features, but the failed branch of
\eqref{eq:combined_reward} overrides their sum with zero.

\begin{table}[!htbp]
  \centering
  \caption{Additive terms in the Lean-based heuristic. Agent execution
  refers to the proving-agent invocation for the current expansion.}
  \label{tab:lean_reward_terms}
  \begin{tabular}{@{}p{0.63\linewidth}l@{}}
    \toprule
    Condition & Contribution to $h$ \\
    \midrule
    Candidate compiles ($b=1$) & $+0.35$ \\
    Proof text changed ($e=1$) & $+0.10$ \\
    Proof text unchanged ($e=0$) & $-0.12$ \\
    Change in unfinished-marker count & $\max\{-0.30,\min\{0.30,0.18\Delta\}\}$ \\
    $m_{u'}>0$ unfinished markers remain & $-\min\{0.20,0.04m_{u'}\}$ \\
    $a>0$ new structured steps & $+\min\{0.20,0.05a\}$ \\
    $a=0$, $e=1$, and $b=1$ & $+0.03$ \\
    Candidate status is \texttt{failed} & $-0.25$ \\
    Agent times out, with / without a text change & $-0.20\;/\;-0.35$ \\
    Agent execution errors without a timeout & $-0.10$ \\
    Either Lean check times out (charged once) & $-0.20$ \\
    Proof text changed but does not compile & $+0.05$ \\
    \bottomrule
  \end{tabular}
\end{table}

\paragraph{LLM progress score.}
For an open candidate, the evaluator receives the target theorem, the
parent and candidate proofs, and the candidate's Lean checking result.
It assesses the change in the proof rather than the absolute quality
of the candidate. The prompt is given below; the placeholders are filled
with the corresponding theorem, proof texts, and checking output.
The numeric \texttt{score} is parsed and checked to lie in $[0,1]$
before combining it with $r^{\mathrm{Lean}}$. The heuristic's numeric
value is not included in the prompt.

\begin{tcolorbox}[
  enhanced,
  title={Prompt Template for LLM Progress Evaluation},
  fonttitle=\bfseries\ttfamily,
  coltitle=white,
  colbacktitle=black!52,
  colback=black!2,
  colframe=blue!45!black,
  boxrule=0.6pt,
  arc=2.5mm,
  left=9pt,right=9pt,top=7pt,bottom=7pt,
  toptitle=4pt,bottomtitle=4pt,
  before skip=10pt,after skip=10pt
]
\textbf{Task:}\par
You are evaluating one step in a Lean proof search. Compare the child
proof with the parent proof for the target theorem. Score the substantive
progress made in this step, not the child's absolute quality.

\medskip
\textbf{Instructions:}\par
Give credit for resolving proof obligations or adding useful intermediate
results. Do not reward longer text, cosmetic edits, unsupported claims,
or removing \texttt{sorry} without a valid proof. Use the Lean result to
check whether claimed progress is credible.

\medskip
\textbf{Scoring:}\par
Score from 0 to 1:\par
\begin{tabular}{@{}ll@{}}
  0 & no progress or regression; \\
  0.25 & small credible step; \\
  0.5 & substantial step; \\
  0.75 & most remaining work resolved; \\
  1 & theorem completely proved.
\end{tabular}\par
When evidence is unclear, score conservatively.

\medskip
\textbf{Input:}\par
\textbf{Target theorem:} \texttt{\{theorem\_statement\}}\par
\textbf{Parent proof:} \texttt{\{parent\_content\}}\par
\textbf{Child proof:} \texttt{\{child\_content\}}\par
\textbf{Child Lean result:} \texttt{\{lean\_result\}}

\medskip
\textbf{Output format:}\par
Return only JSON:\par
\texttt{\{"score": 0.0, "reason": "one concise sentence"\}}
\end{tcolorbox}

\paragraph{Stagnation updates.}
Only a direct attempt to expand $u$ updates $S(u)$; reward backpropagation
does not change the stagnation counts of its ancestors. Let $M(u,u')$
be true if the candidate solves the theorem, or if its text changes and
at least one of the following holds: $\Delta>0$; it compiles and adds
a structured step; or it compiles and has status \texttt{open}.
Thus, the marker-reduction condition does not require compilation,
while a changed, compilable open candidate satisfies this progress test
even without reducing markers or adding structured steps.
Using the final reward $r(u')$ and the parent's mean $Q_{\mathrm{old}}(u)$
before backpropagation, we update
\begin{equation}
 S(u)\leftarrow
 \begin{cases}
  0, & \text{if }M(u,u')\text{ or }r(u')>Q_{\mathrm{old}}(u)+0.03,\\
  S(u)+1, & \text{otherwise}.
 \end{cases}
 \label{eq:stagnation_update}
\end{equation}
This local count determines both the widening slack and the bounded
selection penalty. The next search iteration uses the updated counts
and backed-up rewards.

\FloatBarrier

\section{Details of Symbolic Proving}
\label{app:symbolic_proving_details}

This appendix details the three symbolic proving backends introduced in
Section~\ref{sec:symbolic_proving}. We consider polynomial nonnegativity
goals of the form in Eq.~\ref{eq:standard_polynomial_problem}, with
$p,g_i,h_j\in\mathbb{Q}[\mathbf{x}]$ and variables ranging over
$\mathbb{R}^{n}$. For box-constrained goals, the interval endpoints are
also rational. Each backend searches for an exact certificate for a
polynomial goal supplied by the agent, either directly from the input
theorem or through an intermediate lemma or reformulation. Successful
certificates are translated into local Lean proofs, and unsuccessful
attempts return feedback for further agentic reasoning.

\subsection{SOS-Based Proving under Polynomial Constraints}
\label{app:sos_details}

\paragraph{Degree-bounded SDP formulation.}
We use the polynomial goal and domain $K$ from
Section~\ref{sec:symbolic_proving}. The backend seeks a constrained SOS
identity
\begin{equation}
  p=\sigma_0+\sum_{i=1}^{m}g_i\sigma_i
    +\sum_{j=1}^{\ell}h_jq_j,
  \label{app:sos:certificate}
\end{equation}
where each $\sigma_i$ is a sum of squares and each $q_j$ is a polynomial
multiplier unrestricted in sign. This certificate format is motivated
by Putinar's Positivstellensatz~\citep{putinar1993positive}.
The theorem's existence guarantee requires strict positivity and an
Archimedean quadratic module, modulo the equality ideal when equalities
are present. Our use of the certificate relies on its sufficiency:
any exact identity of this form proves $p\geq0$ on $K$.

Set $g_0=1$ and omit identically zero constraint polynomials.
Choose a relaxation order $d$ such that
$\deg p,\deg g_i,\deg h_j\leq2d$, and define
\begin{equation}
  d_i=\left\lfloor\frac{2d-\deg g_i}{2}\right\rfloor,
  \qquad e_j=2d-\deg h_j.
  \label{app:sos:degree_bounds}
\end{equation}
Let $z_i$ contain all monomials of total degree at most $d_i$ in a
fixed order. We represent $\sigma_i=z_i^{\mathsf T}G_i z_i$ with
$G_i\succeq0$ and impose $\deg q_j\leq e_j$. The resulting identity is
\begin{equation}
  p=\sum_{i=0}^{m}g_i z_i^{\mathsf T}G_i z_i
    +\sum_{j=1}^{\ell}h_jq_j,
  \qquad G_i\succeq0.
  \label{app:sos:truncated_certificate}
\end{equation}
Every summand has degree at most $2d$.
Let $\operatorname{coeff}_{2d}$ extract coefficients in the monomial
basis of total degree at most $2d$, and let $\mathcal A_d(G,q)$ be the
coefficient vector of the right-hand side. Coefficient matching gives
the SDP feasibility problem
\begin{equation}
  \begin{aligned}
    &\text{find }G_0,\ldots,G_m,q_1,\ldots,q_\ell\quad\text{such that}\\
    &\mathcal A_d(G,q)=\operatorname{coeff}_{2d}(p),
      \qquad G_i\succeq0\quad(i=0,\ldots,m).
  \end{aligned}
  \label{app:sos:sdp}
\end{equation}
Here the decision variables are the independent Gram-matrix entries
and the coefficients of the $q_j$. MOSEK supplies numerical candidates
$G_b^{\mathrm{num}}$, where $b=0$ indexes the main SOS block and
$b=1,\ldots,m$ indexes the SOS multipliers of the inequality constraints.

\paragraph{Rational recovery of SOS multipliers.}
We extend the numerical--symbolic certification procedure of
\citet{kaltofen2012exact}, developed for unconstrained problems,
to the constrained SOS identity in Eq.~\ref{app:sos:certificate}.
We recover the certificate sequentially, first fixing the SOS
multipliers $\sigma_1,\ldots,\sigma_m$ and then recovering the
main SOS block together with the equality multipliers.
For a candidate denominator $D$, numerical matrix entries are
rounded to multiples of $1/D$ and symmetrized in exact arithmetic.
Each rational candidate is checked by an exact
$LDL^{\mathsf T}$ decomposition for positive semidefiniteness.
The denominator is adjusted if a candidate fails this check.

For singular blocks with an available exact rational kernel subspace,
recovery can be performed in reduced coordinates. Let the columns of
$P_b$ form a rational basis for the orthogonal complement of that
subspace, and define
\begin{equation}
  L_b=(P_b^{\mathsf T}P_b)^{-1}P_b^{\mathsf T},
  \qquad
  S_b^{\mathrm{num}}=L_bG_b^{\mathrm{num}}L_b^{\mathsf T}.
  \label{app:sos:reduced_coordinates}
\end{equation}
Rational recovery and the exact positivity check are then applied to
$S_b^{\mathrm{num}}$. An accepted $S_b^{\mathbb Q}\succeq0$ gives
$G_b^{\mathbb Q}=P_bS_b^{\mathbb Q}P_b^{\mathsf T}\succeq0$,
preserving the prescribed kernel subspace. This reduction is used
only when the required exact kernel information is available.

After recovering the inequality-multiplier blocks, set
\begin{equation}
  \sigma_i=z_i^{\mathsf T}G_i^{\mathbb Q}z_i\quad(i=1,\ldots,m),
  \qquad
  C_{\mathrm{sm}}:=\sum_{i=1}^{m}g_i\sigma_i,
  \qquad R_0:=p-C_{\mathrm{sm}}.
  \label{app:sos:fixed_multipliers}
\end{equation}
These rational SOS multipliers remain fixed throughout the subsequent
recovery. The remaining task is
\begin{equation}
  R_0=\sigma_0+\sum_{j=1}^{\ell}h_jq_j.
  \label{app:sos:remaining_problem}
\end{equation}
Thus only one SOS block remains to be recovered; the equality
multipliers are still to be determined. With no equality constraints,
the task reduces to an SOS decomposition of $R_0$.

\paragraph{Main-block recovery and exact verification.}
For the remaining main block, we use Gauss--Newton refinement
followed by rational recovery.
Write $\bar z_0=z_0$ without reduction, or
$\bar z_0=P_0^{\mathsf T}z_0$ when using reduced coordinates.
Factor $G_0^{\mathrm{num}}$, or $S_0^{\mathrm{num}}$ in reduced
coordinates, approximately as $U^{\mathsf T}U$.
If $\theta$ collects the entries of $U$ and the coefficients of the
equality multipliers, define
\begin{equation}
  \mathcal R(\theta)=\operatorname{coeff}_{2d}\!\left(
    R_0-\|U\bar z_0\|_2^2-\sum_{j=1}^{\ell}h_jq_j
  \right).
  \label{app:sos:residual}
\end{equation}
At iteration $k$, the Gauss--Newton update solves the linearized
least-squares problem
\begin{equation}
  \begin{aligned}
    \Delta\theta_k
      &\in\underset{\Delta\theta}{\operatorname{argmin}}
        \bigl\|\mathcal R(\theta_k)
        +J_{\mathcal R}(\theta_k)\Delta\theta\bigr\|_2^2,\\
    \theta_{k+1}&=\theta_k+\Delta\theta_k,
  \end{aligned}
  \label{app:sos:gn_step}
\end{equation}
where $J_{\mathcal R}$ is the residual Jacobian. The previously recovered
SOS multipliers enter only through the fixed polynomial $R_0$.

From the refined matrix $U^{\mathsf T}U$, rational reconstruction and
exact coefficient correction seek a main Gram matrix
$G_0^{\mathbb Q}$ and rational equality multipliers satisfying
\begin{equation}
  R_0=z_0^{\mathsf T}G_0^{\mathbb Q}z_0
        +\sum_{j=1}^{\ell}h_jq_j,
  \qquad G_0^{\mathbb Q}\succeq0,
  \qquad q_j\in\mathbb Q[\mathbf{x}].
  \label{app:sos:exact_checks}
\end{equation}
In reduced coordinates, the recovered matrix is $S_0^{\mathbb Q}$ and
$G_0^{\mathbb Q}=P_0S_0^{\mathbb Q}P_0^{\mathsf T}$.
Recovery targets the Gram matrix; the numerical factor $U$ need not
itself have rational entries. The coefficient correction enforces the
identity in Eq.~\ref{app:sos:exact_checks}, and positive
semidefiniteness is checked again after correction.
For a single affine equality $h=0$, the equality multiplier can instead
be obtained by exact division of $R_0-\sigma_0$ by $h$, with zero
remainder required. If recovery or verification fails, recovery
parameters are adjusted and the main block is refined or recomputed,
while the accepted SOS multipliers remain fixed. Exhausting the
available budget returns an unsolved result.

Finally, the complete identity in Eq.~\ref{app:sos:certificate} is
verified coefficientwise over $\mathbb Q$. Exact
$LDL^{\mathsf T}$ decompositions supply weighted-square representations
\begin{equation}
  \sigma_b=\sum_k w_{bk}s_{bk}(\mathbf{x})^2,
  \qquad w_{bk}\in\mathbb Q_{\geq0},\quad
  s_{bk}\in\mathbb Q[\mathbf{x}].
  \label{app:sos:weighted_squares}
\end{equation}
Together with the rational equality multipliers, these representations
form the certificate returned for Lean proof generation. On $K$, all
SOS terms and their products with $g_i$ are nonnegative, while every
$h_jq_j$ vanishes, establishing $p\geq0$.

\subsection{Bernstein-Based Proving under Box Constraints}
\label{app:bernstein_details}

This backend uses Bernstein representations and recursive subdivision
to certify polynomial nonnegativity on a box, following
\citet{ray2009range}. The numerical candidate search and exact recovery
procedure are described below.

\paragraph{Coefficient matching on a box.}
Let $K=\prod_{i=1}^{n}[a_i,b_i]$, with rational endpoints $a_i<b_i$.
The affine map
$T_K(\mathbf y)_i=a_i+(b_i-a_i)y_i$ maps $[0,1]^n$ onto $K$.
Write $\widetilde p_K=p\circ T_K$, and choose a multidegree
$\mathbf d=(d_1,\ldots,d_n)$ with
$d_i\geq\deg_{y_i}\widetilde p_K$. Define
$I_{\mathbf d}=\{\alpha\in\mathbb Z_{\geq0}^{n}:\alpha_i\leq d_i\}$.
The Bernstein basis polynomials in Eq.~\ref{eq:bernstein_basis}
are nonnegative on $[0,1]^n$. Thus an expansion
\begin{equation}
  \widetilde p_K(\mathbf y)
  =\sum_{\alpha\in I_{\mathbf d}}
    c_\alpha B_{\mathbf d,\alpha}(\mathbf y),
  \qquad c_\alpha\geq0,
  \label{app:bernstein:certificate}
\end{equation}
certifies $p\geq0$ on $K$.

Write $\widetilde p_K=\sum_{\beta\in I_{\mathbf d}}
v_\beta\mathbf y^\beta$, including zero coefficients where necessary.
Let $A$ contain the monomial coefficient vectors of the Bernstein
polynomials as its columns. Its entries are
\begin{equation}
  A_{\beta,\alpha}=
  \begin{cases}
    \displaystyle\prod_{i=1}^{n}
      \binom{d_i}{\alpha_i}
      \binom{d_i-\alpha_i}{\beta_i-\alpha_i}
      (-1)^{\beta_i-\alpha_i},
      & \alpha\leq\beta,\\[4pt]
    0, & \text{otherwise},
  \end{cases}
  \label{app:bernstein:matrix}
\end{equation}
where multi-index inequalities are componentwise and both indices
range over $I_{\mathbf d}$. Matching coefficients gives the LP
feasibility problem
\begin{equation}
  \text{find }\mathbf c
  \quad\text{subject to}\quad
  A\mathbf c=\mathbf v,\qquad\mathbf c\geq0.
  \label{app:bernstein:lp}
\end{equation}
For a fixed multidegree, the Bernstein coefficients are unique, so
this LP tests whether the unique coefficient vector is nonnegative.

\paragraph{Numerical candidates and exact recovery.}
We first solve Eq.~\ref{app:bernstein:lp} in floating-point
arithmetic. Given a numerical candidate $\widehat{\mathbf c}$ and a
zero threshold $\tau>0$, we form the tentative support
$S=\{\alpha\in I_{\mathbf d}:|\widehat c_\alpha|>\tau\}$.
Fixing the remaining coefficients to zero gives the reduced problem
\begin{equation}
  \text{find }\mathbf u\in\mathbb Q^{|S|}
  \quad\text{subject to}\quad
  A_{:,S}\mathbf u=\mathbf v,\qquad\mathbf u\geq0,
  \label{app:bernstein:reduced_lp}
\end{equation}
which we solve in exact arithmetic. Here $A_{:,S}$ contains the
columns indexed by $S$. We reconstruct the full vector
$\mathbf c^{\mathbb Q}$ by assigning the recovered coefficients to
$S$ and zero to its complement, then verify
\begin{equation}
  A\mathbf c^{\mathbb Q}=\mathbf v,
  \qquad \mathbf c^{\mathbb Q}\geq0
  \label{app:bernstein:exact_checks}
\end{equation}
exactly. Only a candidate passing both checks certifies the current
box. The numerical threshold proposes a support; it does not establish
that the discarded coefficients are exactly zero. If exact recovery
or verification fails, the attempt returns an unresolved result to
the search controller.

\paragraph{Binary subdivision and certificate construction.}
If no suitable numerical candidate is obtained and the subdivision
budget permits, we choose a coordinate $k$ and a rational split point
$s\in(a_k,b_k)$, and form
\begin{equation}
  K^-=K\cap\{\mathbf x:x_k\leq s\},
  \qquad
  K^+=K\cap\{\mathbf x:x_k\geq s\}.
  \label{app:bernstein:split}
\end{equation}
This replaces $[a_k,b_k]$ by $[a_k,s]$ and $[s,b_k]$, leaving all
other coordinate intervals unchanged. Each child box is normalized
separately, and the same certificate search is applied recursively
to the original polynomial $p$ on that box. Since $K=K^-\cup K^+$,
certificates for both children certify the parent; the shared boundary
is included in both closed boxes.

A successful certificate consists of the binary subdivision tree and
an exact nonnegative Bernstein expansion for every leaf box. The
Translator uses these data to verify the affine substitutions, local
polynomial identities, coefficient nonnegativity, and coverage of the
original box in Lean. If the search budget is exhausted while any
leaf remains unresolved, the backend returns an unsolved result.

\subsection{Successive Difference Substitution}
\label{app:diffsub_details}

DiffSub searches for a finite certificate of polynomial nonnegativity
on $\mathbb R_{\geq0}^{n}$ using successive difference substitution
\citep[Chapter~10]{xia2016automated}. The procedure combines variable
ordering with linear substitutions and certifies branches whose
transformed polynomials have nonnegative coefficients.

\paragraph{Ordering domains and substitutions.}
Let $\mathfrak S_n$ denote the permutations of $\{1,\ldots,n\}$.
For each $\pi\in\mathfrak S_n$, define
\begin{equation}
  C_\pi=\{\mathbf x\in\mathbb R_{\geq0}^{n}\mid
  x_{\pi(1)}\geq\cdots\geq x_{\pi(n)}\}.
  \label{app:diffsub:ordering}
\end{equation}
Every nonnegative tuple admits such an ordering, so
$\bigcup_{\pi\in\mathfrak S_n}C_\pi=\mathbb R_{\geq0}^{n}$.
On each ordering domain, introduce nonnegative difference variables
through the linear map
\begin{equation}
  (T_\pi\mathbf t)_{\pi(i)}=\sum_{j=i}^{n}t_j,
  \qquad i=1,\ldots,n,\qquad\mathbf t\geq0.
  \label{app:diffsub:map}
\end{equation}
This map is a bijection from $\mathbb R_{\geq0}^{n}$ onto $C_\pi$,
with inverse
\begin{equation}
  t_i=x_{\pi(i)}-x_{\pi(i+1)}\quad(i<n),
  \qquad t_n=x_{\pi(n)}.
  \label{app:diffsub:inverse}
\end{equation}
Consequently, a polynomial $q$ is nonnegative on $C_\pi$ if and only
if $q\circ T_\pi$ is nonnegative on $\mathbb R_{\geq0}^{n}$.
The ordering domains may overlap when coordinates are equal, and
their union still covers the entire domain.

\paragraph{Recursive certificate construction.}
The root contains the input polynomial $p$. At a node with polynomial
$q(\mathbf u)$, we expand it exactly and test whether
\begin{equation}
  q(\mathbf u)=\sum_\alpha a_\alpha\mathbf u^\alpha,
  \qquad a_\alpha\geq0\text{ for every }\alpha.
  \label{app:diffsub:leaf}
\end{equation}
Since every monomial is nonnegative for $\mathbf u\geq0$, this
condition certifies the node as a leaf. If the test fails and the
search budget permits, we enumerate all orderings of the current
variables and form the child polynomials
$q_\pi(\mathbf t)=q(T_\pi\mathbf t)$.
Each child passing the coefficient test is closed; unresolved children
are processed recursively by ordering and substituting their own
nonnegative variables. An internal node is certified only when every
ordering branch has been certified.

A successful certificate is a finite tree recording the ordering
cases and substitution identities at internal nodes, together with
the exact polynomial expansions and nonnegative coefficients at the
leaves. The substitutions establish nonnegativity on each ordering
domain, and coverage of these domains establishes nonnegativity at
the parent. Applying this argument recursively proves $p\geq0$ on
$\mathbb R_{\geq0}^{n}$. All substitutions, expansions, and coefficient
checks use exact arithmetic. If the shared search budget is exhausted
while any branch remains unresolved, the backend returns an unsolved
result.

\subsection{Lean Proof Generation and Verification}
\label{app:lean_proof_generation}

The Translator converts exact certificates from the three backends
into local Lean proofs of the supplied polynomial statements. For
SOS, it instantiates the recovered rational weighted-square identity,
proves nonnegativity of the weights and squared terms, and applies
the inequality and equality assumptions to the multiplier terms.
For Bernstein certificates, it formalizes the affine substitutions,
derives the unit-box bounds, and verifies the nonnegative basis
expansion on each leaf box. It then combines these local results
using the subdivision tree and coverage of the parent boxes.
For DiffSub, it constructs the ordering cases, proves nonnegativity
of the difference variables, and verifies the substitution identities
and coefficient nonnegativity at the leaves. Polynomial identities
and rational arithmetic are verified within Lean. A local goal is
discharged only after the generated proof is accepted by the Lean
kernel.

The agent incorporates the verified local result into the surrounding
proof context and establishes any conditions connecting it to the
original goal. These include the assumptions required by the local
certificate and the side conditions of any reformulation, such as
denominator positivity or variable-substitution identities.
The resulting complete proof is checked against the original theorem
statement, with its original assumptions, by the Lean kernel.

\section{Sources and Construction of LitIneq}
\label{app:research_benchmark_sources}
We construct LitIneq by collecting 74 examples from research papers
and formal-verification projects and formulating them as
inequality-proving tasks in Lean~4.
The benchmark combines polynomial bound-verification problems
from global optimization with nonlinear inequalities from
geometric and program verification.

\paragraph{Source collection.}
Table~\ref{tab:research_sources} summarizes the composition.
We collect 41 polynomial instances from studies on Bernstein
polynomials, global optimization, and inequality
verification~\citep{munoz2013bernstein,liu2025appirl,ray2009range},
together with examples from the Flyspeck nonlinear-verification
documentation. \footnote{\href{https://github.com/flyspeck/flyspeck/blob/master/formal_ineqs/docs/FormalVerifier.tex}
{\emph{A Tool for Formal Verification of Nonlinear Inequalities}},
Alexey Solovyev, Flyspeck project documentation.}
For C1--C10, we retain the APPIRL benchmark formulations and identifiers. Table~\ref{tab:research_sources} lists their original sources as attributed in APPIRL.

\begin{table}[!htbp]
  \caption{Collection sources and source attributions for the 74 LitIneq instances. C1--C10 attributions follow APPIRL.}
  \label{tab:research_sources}
  \centering
  \setlength{\tabcolsep}{4pt}
  \begin{tabular}{@{}p{0.25\linewidth}rp{0.64\linewidth}@{}}
    \toprule
    Source group & Count & Instances and source attributions \\
    \midrule
    Bernstein examples & 2
      & Coefficient and subdivision examples \\
    APPIRL benchmark & 2
        & C1--C2: Verschelde (1996), \emph{Polynomial homotopy continuation:
          A portable Ada software package}~\citep{verschelde1996phc} \\
      APPIRL benchmark & 1
        & C3: Iserles (1995), personal communication at the AMS--SIAM
          summer seminar~\citep{iserles1995personal} \\
      APPIRL benchmark & 2
        & C4, C6: Noonburg (1989), \emph{A neural network modeled by an
          adaptive Lotka--Volterra system}~\citep{noonburg1989neural} \\
      APPIRL benchmark & 1
        & C5: Verschelde (2001), \emph{The PHC Pack, the database of
          polynomial systems}~\citep{verschelde2001phc} \\
      APPIRL benchmark & 1
        & C7: Faug\`ere (1999), \emph{A new efficient algorithm for
          computing Gr\"obner bases (F4)}~\citep{faugere1999new} \\
      APPIRL benchmark & 1
        & C8: Wright (1985), \emph{Finding all solutions to a system of
          polynomial equations}~\citep{wright1985finding} \\
      APPIRL benchmark & 1
        & C9: Emiris (1994), \emph{Sparse elimination and applications in
          kinematics}~\citep{emiris1994sparse} \\
      APPIRL benchmark & 1
        & C10: Emiris (1997), \emph{A general solver based on sparse
          resultants: Numerical issues and kinematic applications}
          ~\citep{emiris1997general} \\
    Polynomial bounds from verification examples & 8
      & Reaction Diffusion, Schwefel, Lotka--Volterra, Magnetism,
        Caprasse, Flyspeck Delta, Butcher, and Heart Dipole \\
    Bernstein range-computation benchmark & 21
      & A02, A04--A07, A09, A11--A18, and B01--B07 \\
Flyspeck (HOL Light) & 11 & Original example identifiers \\
NASALib (PVS) & 22 & Original lemma names \\
    \midrule
    Total & 74 & \\
    \bottomrule
  \end{tabular}
\end{table}
The remaining 33 instances come from Flyspeck's HOL Light
development\footnote{\url{https://github.com/flyspeck/flyspeck}}
and NASA's PVS library,
NASALib.\footnote{\url{https://github.com/nasa/pvslib}}

\paragraph{Task construction and Lean formalization.}
For the 41 polynomial instances, we formulate bound-verification
problems over explicit box domains. A selected lower bound
$p(\mathbf{x})\geq\ell$ is expressed as
$p(\mathbf{x})-\ell\geq0$, with upper bounds handled analogously.
Finite decimal coefficients and bounds are represented as exact
rational numbers. For the HOL Light and PVS examples, we translate the selected
statements and required definitions into Lean~4, making domain
restrictions and positivity assumptions explicit.
Each instance is packaged as a separate theorem with its own
assumptions and target, retaining the strict or non-strict relation
of the selected statement. Source identifiers and statement
transformations are recorded for each instance.

Some instances involve explicit choices of domain or target.
For A14 (Magnetism), we specify the box $[-5,5]^7$ to resolve
the ambiguous domain specification in the source.
For B02 (Six-hump camel), we construct the target by subtracting
a decimal approximation of the reported minimum from the
objective polynomial.
For B05 (Quadratic), we take $\sum_i x_i^2$ as the nonnegative
target. For Flyspeck's Delta bound, the 41-instance bound-verification
collection contains a non-strict version, while the HOL Light
collection retains the strict version.

\paragraph{Polynomial instances.}
Of the 74 instances, 49 have targets and assumptions consisting
of polynomial comparisons with rational coefficients.
These comprise the 41 polynomial instances described above,
five Flyspeck instances, and three NASALib instances.
Table~\ref{tab:litineq_polynomial_profile} summarizes their
distribution by the number of quantified real variables and
the total degree of the target polynomial.
Forty-six instances have box constraints, with two bound inequalities
per variable and no explicit equality constraints.
Among the remaining three, \texttt{simple} is unconstrained,
\texttt{quadratic} has two polynomial inequality assumptions,
and \texttt{ajn} has two alternative branches, each containing
one equality and eleven inequalities.

\begin{table}[!htbp]
  \centering
  \caption{Distribution of the 49 polynomial instances by the number
  of variables $n$ and target total degree $d_p$. Entries are instance
  counts. We count all quantified real variables, including parameters,
  without eliminating fixed variables. The degree is computed after
  expanding auxiliary definitions, simplifying rational constants,
  and moving the target to one side.}
  \label{tab:litineq_polynomial_profile}
  \begingroup
  \small
  \setlength{\tabcolsep}{9pt}
  \renewcommand{\arraystretch}{1.1}
  \begin{tabular}{@{}lrrrr@{}}
    \toprule
    & \multicolumn{3}{c}{Target total degree $d_p$} & \\
    \cmidrule(lr){2-4}
    Variables $n$ & $0$--$2$ & $3$--$4$ & $5$--$8$ & Total \\
    \midrule
    $1$--$2$ & 6 & 2 & 2 & 10 \\
    $3$--$4$ & 4 & 9 & 0 & 13 \\
    $5$--$6$ & 7 & 7 & 3 & 17 \\
    $7$--$8$ & 4 & 2 & 2 & 8 \\
    $11$ & 0 & 1 & 0 & 1 \\
    \midrule
    Total & 21 & 21 & 7 & 49 \\
    \bottomrule
  \end{tabular}
  \endgroup
\end{table}

\paragraph{Instances involving non-polynomial expressions.}
The remaining 25 instances comprise six Flyspeck instances
and nineteen NASALib instances. They involve rational expressions with
variable-dependent denominators, absolute
values, square roots, logarithms, exponentials, and trigonometric
or inverse trigonometric functions, including sine, tangent,
and arctangent.
Among them, 21 statements contain between one and eight quantified
real variables, counting parameters as variables, while four are
closed numerical inequalities.
Non-polynomial expressions may occur in the target, the assumptions,
or the constants defining the domain.
The classification follows the input expressions before
proof-oriented reformulations.

\section{Evaluation Protocol}
\label{app:evaluation_protocol}

All methods are evaluated with a two-hour time limit per problem. We use
Lean and Mathlib v4.30.0 for all Lean-based methods. Each task retains its
original assumptions and target statement; a successful Lean proof must
establish that statement without \texttt{sorry} or \texttt{admit}.

\paragraph{Hardware and agent runtime.}
The local evaluation platform uses an Intel Core i9-10900K CPU at
3.70\,GHz with 64\,GB of DDR4 memory. IneqAgent and Direct Agent use
\texttt{codex-cli} version \texttt{0.149.0} as the underlying LLM coding
agent. API-based LLM inference is served externally; the locally deployed
prover models use the GPUs specified below.

\paragraph{Symbolic computation.}
For Z3, we encode the assumptions together with the negated target as an
SMT satisfiability query. An \texttt{unsat} result establishes the
inequality, whereas \texttt{sat} indicates a counterexample; an
\texttt{unknown} result or timeout is not counted as a success.
For Maple, we use \texttt{PolynomialRing} from the \texttt{RegularChains}
package to specify the polynomial ring and invoke \texttt{SamplePoints}
to determine whether the corresponding counterexample constraints have
real solutions. A problem is counted as solved when the computation
establishes that no such solution exists. For Isabelle/HOL, we apply the
\texttt{sos} method and count a success when Isabelle completes the
formal proof.

\paragraph{LLM provers.}
We evaluate Goedel-Prover-V2-8B and DeepSeek-Prover-V2-7B at pass@32.
A problem is solved if at least one of up to 32 proof attempts produces
a Lean-verified proof within the two-hour limit. Both models are deployed
for inference on NVIDIA L40S GPUs.

\paragraph{Agentic frameworks.}
For the published agentic baselines, we use the configurations associated
with the best reported results in their respective papers, subject to
the common two-hour limit. Our framework uses DeepSeek-V4-Flash for
the main comparison in Table~\ref{tab:main_results}. Agentic methods have
no additional caps on proof attempts, LLM calls, or token usage, and API
inference is provided by external services. In the cost comparison in
Table~\ref{tab:cost_comparison}, our framework and Direct Agent use the
same backbone in each pair. Direct Agent runs with its default
configuration and access to Lean.

We report the percentage of problems solved and the mean time in seconds
over successful problems. We compute overall success rate over all 186
problems and overall mean time over all problems successfully
solved by each method across the four datasets.

\paragraph{LLM inference pricing.}
Table~\ref{tab:llm_api_pricing} lists the token rates used for the four
backbones. Uncached input, cached input, and output tokens are charged
at their respective rates, and inference cost is accumulated over all
LLM calls in a run. Each IneqAgent--Direct Agent pair uses the same
backbone and rates.  Rates are reported in USD per million tokens and the cost comparisons
in Section~\ref{sec:cost_comparison} are reported in US cents.
\begin{table}[!htbp]
  \centering
  \caption{LLM inference rates in USD per million tokens. Input denotes uncached input tokens.}
  \label{tab:llm_api_pricing}
  \begin{tabular}{@{}llrrr@{}}
    \toprule
    Backbone & Input & Cached input & Output \\
    \midrule
    DeepSeek-V4-Flash & 0.2206 & 0.0074 & 0.6618 \\
    Qwen3.8-Flash & 0.1176 & 0.0147 & 0.3971 \\
    Kimi-K3 & 2.9412 & 0.2941 & 14.7059 \\
    GPT-5.5 & 5.0000 & 0.5000 & 30.0000 \\
    \bottomrule
  \end{tabular}
\end{table}
\FloatBarrier

\section{Details of the Dittert, Holens--Djokovic, and Merris Conjectures}
\label{app:conjecture_proofs}

We give the historical context of the conjectures in Section~4.5, the
Dittert certificate constructions, and the proof of our
four-dimensional Merris result. Throughout, $J_n$ denotes the matrix
whose entries are $1/n$, $\Omega_n$ is the set of doubly stochastic
$n\times n$ matrices, and $A(i\mid j)$ is obtained by deleting row $i$
and column $j$. We use $A\geq0$ for entrywise nonnegativity and
$Q\succeq0$ for positive semidefiniteness of a symmetric matrix.

Table~\ref{tab:conjecture_complexity} shows that Dittert $n=5$ has the
largest expanded target, while HD $(4,4)$ incurs the highest time and
cost.

\begin{table}[!htbp]
  \caption{Conjecture task sizes and costs.
  Dittert sizes refer to $P_4$ and $\widetilde F_5$ before constraint elimination.
  For HD $(4,4)$, sizes refer to the reduced Holens--Djokovic polynomial $Q_4$,
  using the upper-left $3\times3$ entries as nine independent variables.
  Time is wall-clock time and Cost is total LLM inference cost.}
  \label{tab:conjecture_complexity}
  \centering
  \small
  \begin{tabular}{@{}lrrrrr@{}}
    \toprule
    Task & Variables & Total degree & Monomials & Time (s) & Cost (US cents) \\
    \midrule
    Dittert, $n=4$  & 16 & 4 & 521 & 5,292 & 84.6728 \\
    Dittert, $n=5$ & 25 & 5 & 118,755 & 6,704 & 100.1654 \\
    HD $(4,4)$ & 9 & 4 & 222 & 42,731 & 521.9610 \\
    \bottomrule
  \end{tabular}
\end{table}

\subsection{Dittert's Conjecture}
\label{app:dittert}

Let $K_n=\{A\in\mathbb{R}_{\geq0}^{n\times n}:\sum_{i,j}a_{ij}=n\}$,
and let $r_i$ and $c_j$ be the row and column sums of $A$.
Dittert's conjecture, recorded in Minc's 1983 survey and restated as
Conjecture~28 in the survey of Cheon and
Wanless~\citep{minc1983theory,cheon2005survey}, asserts
\begin{equation}
 \Phi_n(A):=\prod_{i=1}^{n}r_i+\prod_{j=1}^{n}c_j-\operatorname{per}(A)
 \leq 2-\frac{n!}{n^n}\qquad(A\in K_n),
 \label{eq:app-dittert}
\end{equation}
with equality exactly at $J_n$. On $\Omega_n$, both products of sums
equal one, so the inequality reduces to the van der Waerden lower bound
$\operatorname{per}(A)\geq n!/n^n$. The additional difficulty is to
control matrices in the larger simplex $K_n$, where neither the row sums
nor the column sums are fixed individually.

Sinkhorn proved the $n=2$ case and showed that a maximizing matrix has
positive permanent~\citep{sinkhorn1984dittert}. Hwang established that
any entrywise positive maximizer must be $J_n$ and that $J_n$ is a strict
local maximum~\citep{hwang1986permanents}; he subsequently proved the
$n=3$ case~\citep{hwang1987dittert}. These results leave the exclusion of
nonuniform boundary maximizers as a central obstacle. Cheon and Wanless
excluded partly decomposable matrices and matrices whose entire zero
set forms a single block, and derived quantitative restrictions on the
row and column sums of a possible counterexample~\citep{cheon2012dittert}.
Udayan and Somasundaram also established full indecomposability of a
maximizer in dimension four~\citep{udayan2024lih}. These structural
conditions alone do not identify the maximizer in that dimension.
More recently, a 2026 preprint of Pang reports a proof for all
$n\geq17$ using refined scaling estimates~\citep{pang2026dittert}.
Our constructions below address dimensions four and five by explicit
rational nonnegativity certificates. Both results include quantitative
stability and uniqueness of the maximizer.

\subsubsection{Dimension Four: A Quantitative Stability Certificate}
\label{app:dittert-four}

\paragraph{Polynomial target and certificate template.}
Flatten the entries of $A$ into $x=(x_1,\ldots,x_{16})$, and set
\begin{equation}
 h(x)=\sum_{i=1}^{16}x_i-4,\qquad
 \Delta_4(x)=\sum_{i=1}^{16}(x_i-\tfrac14)^2,\qquad
 P_{4,\lambda}(x)=61-32\Phi_4(A(x))-\lambda\Delta_4(x).
 \label{eq:app-d4-target}
\end{equation}
It suffices to prove $P_{4,\lambda}\geq0$ under $x_i\geq0$ and $h=0$
for one rational $\lambda>0$. We search for a degree-four constrained
SOS identity of the form
\begin{equation}
 P_{4,\lambda}
 =\sigma_0+\sum_{i=1}^{16}x_i\sigma_i
   +\sum_{1\leq i<j\leq16}x_ix_j\sigma_{ij}+h q,
 \label{eq:app-d4-certificate}
\end{equation}
where $\sigma_0$ is SOS of degree at most four, $\sigma_i$ and
$\sigma_{ij}$ are SOS of degree at most two, and $\deg q\leq3$.
Every term except $hq$ is nonnegative on the domain, and $hq$ vanishes
there. The initial template using only $x_i\sigma_i$ does not yield a
positive stability parameter in the numerical search. Adding the
pairwise generators $x_ix_j$ yields a numerical value near $0.6153852$,
which motivates the rational candidate $\lambda=8/13$.
This value is accepted only after exact certification; we do not claim
that it is the optimal stability parameter.

The resulting template has 136 small Gram blocks, corresponding to
$16+\binom{16}{2}$ generators, and one main block. The small blocks use
$z_{\mathrm{sm}}=(1,x_1,\ldots,x_{16})^\top$ and have size $17\times17$, and 
the main block uses all monomials of degree at most two and has size
$153\times153$. A numerical semidefinite solve supplies candidate
matrices. Turning them into exact certificates requires handling both
singularity and the coupling between blocks.

\paragraph{Removing the forced singular direction.}
Write $c=(1/4,\ldots,1/4)$. Since $P_{4,8/13}(c)=h(c)=0$ and every
generator $x_i,x_ix_j$ is positive at $c$, each SOS block must vanish at
$c$. Thus, for a block $\sigma_b=z_b^\top G_bz_b$ with $G_b\succeq0$,
we have $G_bz_b(c)=0$. Ordering $z_b$ with its constant monomial first,
write $z_b(c)=(1,\widetilde z_b(c)^\top)^\top$ and define
\begin{equation}
 R_b=\begin{pmatrix}-\widetilde z_b(c)^\top\\I\end{pmatrix},
 \qquad G_b=R_bS_bR_b^\top.
 \label{eq:app-d4-reduction}
\end{equation}
This parameterization enforces the known kernel exactly. The numerical
reduced matrices $S_b$ are positive definite, making rational recovery
possible without perturbing the forced zero direction.

\paragraph{Sequential rational recovery.}
We recover and fix the 136 small reduced blocks first. Their entries
are rounded to rational values with a common candidate denominator,
and each block is accepted only when an exact $LDL^\top$ factorization
has positive diagonal entries. The successful common denominator in
the construction is 1331. Once these blocks are fixed, let
\[
 C_{\mathrm{sm}}=\sum_i x_i\sigma_i+\sum_{i<j}x_ix_j\sigma_{ij},
 \qquad \mathcal{R}=P_{4,8/13}-C_{\mathrm{sm}}.
\]
The main block is rationalized and corrected to satisfy two conditions:
its reduced matrix remains positive definite, and its residual vanishes
identically after imposing the simplex equality. Explicitly, with
$\sigma_0=z_0^\top R_0S_0R_0^\top z_0$, where
$R_0$ denotes the matrix $R_b$ for the main block, we check
\begin{equation}
 S_0\succ0,\qquad
 (\mathcal{R}-\sigma_0)\left(x_1,\ldots,x_{15},4-\sum_{i=1}^{15}x_i\right)
 \equiv0.
 \label{eq:app-d4-residual}
\end{equation}
The successful denominator parameter for the main-block recovery is
32767. Exact polynomial division then yields $q=(\mathcal{R}-\sigma_0)/h$
with zero remainder, establishing Eq.~\ref{eq:app-d4-certificate}
coefficientwise over $\mathbb{Q}$.

\paragraph{Scaling and the exact code target.}
The normalized target above and the integer-scaled certificate represent
the same inequality. Define
\begin{equation}
 Q_{4,\mathrm{int}}=61-32\Phi_4(A)=32\bigl(61/32-\Phi_4(A)\bigr),
 \qquad D_4^s=\sum_{i=1}^{16}x_i^2-1.
 \label{eq:app-d4-integer-gap}
\end{equation}
With our convention $h=\sum_i x_i-4$, the centered distance satisfies
$\Delta_4=D_4^s-h/2$. Consequently,
\begin{align}
 F_s&:=13Q_{4,\mathrm{int}}-8D_4^s,\notag\\
 F_c&:=13Q_{4,\mathrm{int}}-8\Delta_4
     =F_s+4h=13P_{4,8/13}\notag\\
    &=416\left(\frac{61}{32}-\Phi_4(A)-\frac1{52}\Delta_4\right).
 \label{eq:app-d4-scalings}
\end{align}
The exported SOS identity for $F_s$ has the same block structure as
Eq.~\ref{eq:app-d4-certificate}, with the weights scaled by 13.
If its equality multiplier is $q_{\mathrm{json}}$, the identity for
$F_c$ uses $q_{\mathrm{json}}+4$ and the same SOS blocks.
The Lean definition \texttt{compactTarget} is $F_c$,  dividing its
nonnegativity statement by $416$ gives the stability constant $1/52$.
The coefficient $8/13$ belongs to the 32-fold scaled gap in $P_4$;
it must be divided by 32 exactly once to obtain $1/52$.

\paragraph{Conclusion and formal verification.}
The final certificate contains 152 positively weighted rational squares
in the main SOS and 16 in each of the 136 small SOS blocks. The Lean
formalization checks the rational weights, the nonnegativity of the
constraint factors, and the polynomial identity, and relates the
flattened polynomial to the matrix statement. Therefore
\begin{equation}
 \frac{61}{32}-\Phi_4(A)
 \geq\frac{1}{52}\lVert A-J_4\rVert_F^2\qquad(A\in K_4).
 \label{eq:app-d4-stability}
\end{equation}
If $\Phi_4(A)=61/32$, the squared norm must vanish, so $A=J_4$.
Conversely, direct evaluation at $J_4$ gives equality. This establishes
both the conjectured upper bound and the unique equality case.
The updated proof record identifies \texttt{DH4\_one\_div\_52\_sos}
as the coordinate-level stability theorem and lists matrix theorems
for quantitative stability and the full four-dimensional conjecture
in \texttt{DH4DittertOneDiv52Fast.lean}.

\subsubsection{Dimension Five: Symmetry Reduction and Exact Recovery}
\label{app:dittert-five}

\paragraph{Homogenization, distance term, and symmetry.}
For an arbitrary nonnegative $5\times5$ matrix, put $s=\sum_{i,j}a_{ij}$
and define
\begin{equation}
 \begin{aligned}
 H_5(A)&=1226s^5-5^9\Phi_5(A),\\
 G_5(A)&=s^3\sum_{i,j=1}^{5}(25a_{ij}-s)^2
       =625s^3\sum_{i,j=1}^{5}a_{ij}^2-25s^5,\\
 \widetilde F_5(A)&=100H_5(A)-G_5(A).
 \end{aligned}
 \label{eq:app-d5-target}
\end{equation}
The strengthened certificate matches $\widetilde F_5$, rather than
$H_5$ alone. It is a homogeneous degree-five polynomial in 25 variables
with 118,755 monomials after expansion and collection. The factor $s^3$
makes the squared-distance term homogeneous of degree five as well.
Independent row and column permutations, together with transposition,
form a symmetry group $\mathcal T$
of order $2(5!)^2=28{,}800$. Its action partitions the degree-five
monomials into 54 orbits. Let $v(A)$ list the
$\binom{26}{2}=325$ degree-two monomials in the entries. We seek a
rational symmetric matrix $Q\succeq0$ such that
\begin{equation}
 \widetilde F_5(A)=\mathcal{S}_Q(A)
 :=\sum_{\tau\in\mathcal T}\tau\!\left(a_{11}\,v(A)^\top Qv(A)\right),
 \label{eq:app-d5-certificate}
\end{equation}
where $\tau$ acts by permuting matrix-entry variables. The factor $a_{11}$
raises the degree from four to five and is nonnegative on the domain.
Since both sides are invariant, coefficient matching requires only
one equation per monomial orbit, giving 54 linear equations for $Q$.

\paragraph{Enforcing the equality-point kernel.}
A numerical semidefinite feasibility solve gives an approximate Gram
matrix. At the all-ones matrix, $H_5=G_5=\widetilde F_5=0$ and
$v=\mathbf{1}_{325}$, and  
positive semidefiniteness therefore forces $Q\mathbf{1}_{325}=0$.
To preserve this condition throughout rational recovery, write
$u=\mathbf{1}_{324}$ and parameterize
\begin{equation}
 Q(S)=\begin{pmatrix}S&-Su\\-u^\top S&u^\top Su\end{pmatrix}
     =R^\top S R,\qquad R=(I_{324}\; -u).
 \label{eq:app-d5-completion}
\end{equation}
Here $S$ is a symmetric $324\times324$ matrix. This parameterization
enforces symmetry and the zero direction identically, and
$Q(S)\succeq0$ if and only if $S\succeq0$.

\paragraph{Exact coefficient correction.}
Vectorize the upper triangle of $S$ into
$t\in\mathbb{Q}^{52{,}650}$. The orbit equations become a rational
linear system $Bt=b$, where $b$ contains the orbit coefficients of
$100H_5-G_5$. Strengthening the target changes this right-hand side;
the symmetry operator and kernel parametrization are unchanged.
After rounding the numerical reduced matrix to
a rational candidate $t_{\mathrm{rat}}$, the residual is
$b-Bt_{\mathrm{rat}}$. The completed matrix always has the zero
direction, so evaluation at the all-ones matrix gives one dependency
among the 54 equations. The construction identifies 53 independent
rows and verifies their independence by exact elimination.
Writing the selected system as $B_*t=b_*$, an exact correction is
\begin{equation}
 \delta t=B_*^\top(B_*B_*^\top)^{-1}
                  (b_*-B_*t_{\mathrm{rat}}),\qquad
 t_{\mathrm{exact}}=t_{\mathrm{rat}}+\delta t.
 \label{eq:app-d5-correction}
\end{equation}
This is the minimum Euclidean-norm correction in the chosen upper-triangle
coordinates. It limits the perturbation of the numerical candidate but
does not itself prove positive semidefiniteness. All 54 original equations
are checked again in exact arithmetic after the correction.

\paragraph{Exact positivity and quantitative stability.}
Positive semidefiniteness must be checked in exact arithmetic after
coefficient correction. An exact $LDL^\top$ decomposition of the reduced
matrix with nonnegative diagonal weights suffices: by
Eq.~\ref{eq:app-d5-completion}, it gives a rationally weighted
sum-of-squares representation of $v^\top Qv$.
The strengthened certificate is accepted only after checking symmetry,
the exact kernel equation, positive semidefiniteness, and all coefficient
identities for $100H_5-G_5$.
Each summand in Eq.~\ref{eq:app-d5-certificate} is consequently
nonnegative, proving $100H_5(A)-G_5(A)\geq0$ for every $A\geq0$,
without a normalization assumption. At $s=5$, let
$\Delta_5=\lVert A-J_5\rVert_F^2$. Then
\begin{equation}
 G_5=5^7\Delta_5,\qquad
 \frac{\widetilde F_5}{100\cdot5^9}
 =\frac{1226}{625}-\Phi_5(A)-\frac1{2500}\Delta_5.
 \label{eq:app-d5-scalings}
\end{equation}
Thus the normalized matrix conclusion is
\begin{equation}
 \frac{1226}{625}-\Phi_5(A)
 \geq\frac1{2500}\lVert A-J_5\rVert_F^2\qquad(A\in K_5).
 \label{eq:app-d5-result}
\end{equation}
If $\Phi_5(A)=1226/625$, the positive stability coefficient forces
$\Delta_5=0$, hence $A=J_5$. Direct substitution gives equality at $J_5$.
This establishes the conjectured upper bound and uniqueness, including
on the boundary of $K_5$. Uniqueness follows from the distance estimate;
it does not require the strengthened remainder itself to have only one zero.

\paragraph{Lean verification and the matrix theorem.}
The updated proof record identifies \texttt{h5Poly} with $H_5$,
\texttt{uniquenessGapPoly} with $G_5$, and
\texttt{compactTargetPoly} with $100H_5-G_5$.
The theorem \texttt{eval\_compactTargetPoly\_nonnegative} checks the
strengthened target on the nonnegative orthant; \texttt{DH5\_strengthened}
then states $G_5/100\leq H_5$.
The matrix-level file \texttt{DH5Main.lean} applies $s=5$ and
Eq.~\ref{eq:app-d5-scalings} to obtain the $1/2500$ stability bound
and the unique equality case. These are the quantitative-stability
and dimension-five conjecture theorems listed in the updated record. 

\subsection{Holens--Djokovic (4,4) and Merris's Conjecture}
\label{app:merris}

Merris proposed in 1973 that every $A\in\Omega_n$ satisfies
\begin{equation}
 \min_{1\leq j\leq n}S_j(A)\leq n\operatorname{per}(A),\qquad
 S_j(A):=\sum_{i=1}^{n}\operatorname{per}(A(i\mid j))
 \label{eq:app-merris}
\end{equation}
\citep{merris1973permanent}. The permanent has the weighted column
expansion $\operatorname{per}(A)=\sum_i a_{ij}\operatorname{per}(A(i\mid j))$.
The conjecture asks for at least one column whose corresponding
\emph{unweighted} sum of minors is at most $n$ times the permanent;
this does not follow from the expansion alone.

A related result of Marcus and Merris gives
$H^{1/2}\operatorname{Padj}(H)H^{1/2}\preceq
n\operatorname{per}(H)I_n$ for positive semidefinite Hermitian $H$,
where $\operatorname{Padj}(H)_{ji}=\operatorname{per}(H(i\mid j))$.
For positive semidefinite doubly stochastic matrices, their averaged
minor inequality implies Eq.~\ref{eq:app-merris}
\citep{marcus1973adjoints}. General doubly stochastic matrices need
not be symmetric or positive semidefinite, so this argument does not
cover the conjecture in full. Udayan and Somasundaram's 2021 paper
records the general problem as open for $n\geq4$ and proves it for
additional classes, including matrices constant within each column
whose largest entry is at least $1/n$~\citep{udayan2021merris}.

\subsubsection{Dimension Four: An SOS Proof via Holens--Djokovic}
\label{app:merris-four}

\paragraph{Merris target and the certified polynomial.}
For $A\in\Omega_4$, write
\begin{equation}
 S_j(A)=\sum_{i=1}^{4}\operatorname{per}(A(i\mid j)),\qquad
 M_{4,j}(A)=4\operatorname{per}(A)-S_j(A).
 \label{eq:app-merris-four-target}
\end{equation}
The desired conclusion is
\begin{equation}
 \forall A\in\Omega_4,\qquad
 \exists j\in\{1,2,3,4\},\quad M_{4,j}(A)\geq0.
 \label{eq:app-merris-four-result}
\end{equation}
The SOS certificate directly proves the $(4,4)$ instance of the
Holens--Djokovic conjecture~\citep{holens1964doubly,djokovic1967waerden}
in the polynomial form
\begin{equation}
 Q_4(A):=24\operatorname{per}(A)-2-\sum_{i,j=1}^{4}a_{ij}^{3}\geq0,
 \qquad A\in\Omega_4.
 \label{eq:app-hd4-target}
\end{equation}
As shown below, this implies a bound on the average of the four $S_j$,
which in turn gives Eq.~\eqref{eq:app-merris-four-result}.
Thus $Q_4$ is a sufficient polynomial target for the Merris result, and 
we do not identify it with any individual $M_{4,j}$.

\begin{remark}[On a previous claim concerning the $(4,4)$ case]
\label{rem:hd44_prior_claim}
\citet[Theorem~6]{udayan2021merris} claim that the
Holens--Djokovic inequality fails for $n=k=4$.
Their proof defines
\[
  \Delta(t)
  = \operatorname{per}(A)
    - \operatorname{per}\bigl(tJ_4+(1-t)A\bigr)
\]
and subsequently uses $\Delta(1)=0$.
However, this definition gives
\[
  \Delta(1)
  = \operatorname{per}(A)-\operatorname{per}(J_4)
  = \operatorname{per}(A)-\frac{3}{32},
\]
which is generally nonzero.
For example, $A=I_4$ gives $\Delta(1)=29/32$.
Consequently, the subsequent derivative argument,
which relies on a zero at $t=1$, does not establish
the claimed disproof.
The exact rational SOS certificate presented below
independently establishes
\eqref{eq:hd44_polynomial} for every $A\in\Omega_4$.
\end{remark}

\paragraph{Nine-variable parametrization.}
Let $u=(u_{11},u_{12},\ldots,u_{33})$ contain the upper-left $3\times3$
entries. Define $A(u)$ by
\begin{align}
 a_{ij}&=u_{ij} &&(i,j\leq3),\notag\\
 a_{i4}&=1-\sum_{j=1}^{3}u_{ij},\qquad
 a_{4j}=1-\sum_{i=1}^{3}u_{ij} &&(i,j\leq3),\notag\\
 a_{44}&=\sum_{i,j=1}^{3}u_{ij}-2.
 \label{eq:app-hd4-parametrization}
\end{align}
All row and column sums are then one. Every matrix in $\Omega_4$ has
this representation, and its feasible domain is exactly the sixteen
linear inequalities $a_{ij}(u)\geq0$. Consequently, the parametrization
eliminates the equality constraints without relaxing nonnegativity.
The polynomial $\widehat Q_4(u)=Q_4(A(u))$ has nine variables, total
degree four, and 222 nonzero monomials after expansion and collection.
These are the sizes reported for the Merris experiment in
Section~4.5; 222 counts the terms of this single
SOS target.

\paragraph{Symmetrized rational SOS certificate.}
Let $G$ consist of the row permutations, column permutations, and
transposition, with $|G|=2(4!)^2=1152$. These transformations preserve
$\Omega_4$ and $Q_4$. The certificate uses four representative multipliers,
\begin{equation}
 m_0(A)=a_{41}a_{42}a_{43}a_{44},\quad
 m_1(A)=a_{44},\quad m_2(A)=a_{44}a_{43},\quad m_3(A)=a_{44}a_{33},
 \label{eq:app-hd4-multipliers}
\end{equation}
and has the form
\begin{equation}
 \widehat Q_4(u)=
 \sum_{r=0}^{3}\sum_{\tau\in G}\sum_{k=1}^{N_r}
 \lambda_{r,k}\,m_r(\tau A(u))\,q_{r,k}(\tau A(u))^2,
 \qquad \lambda_{r,k}\in\mathbb{Q}_{\geq0}.
 \label{eq:app-hd4-sos}
\end{equation}
Here each $q_{r,k}$ has rational coefficients; when expressed in nine
coordinates, its argument is the upper-left block of $\tau A(u)$.
The four blocks have $(N_0,N_1,N_2,N_3)=(9,24,42,34)$ weighted squares
before summing over transformations. The implementation partitions the
blocks into $(8,16,24,24)$ slices, respectively, for a total of 72 slices.
This partition permits parallel checking of the same algebraic identity.
For every feasible $u$, all transformed matrix entries, multipliers,
weights, and squares are nonnegative. Hence the certificate establishes
Eq.~\ref{eq:app-hd4-target}, including matrices with zero entries.

The certificate is discovered by semidefinite programming on the
nine-variable model. Known equality cases, including the uniform matrix,
impose vanishing conditions that reduce the Gram-block dimensions.
Numerical Gram data are recovered over the rationals and factored into
weighted squares. The exported weights, polynomials, symmetry mappings,
and target coefficients are then checked by exact polynomial expansion.
The numerical search supplies a candidate; acceptance depends on the
rational identity and nonnegativity checks.

\paragraph{From Holens--Djokovic to Merris.}
For doubly stochastic $4\times4$ matrices, the proof uses the identity
\begin{equation}
 3\sum_{j=1}^{4}S_j(A)=4+2\sum_{i,j=1}^{4}a_{ij}^{3}.
 \label{eq:app-hd4-minor-identity}
\end{equation}
After substitution of Eq.~\ref{eq:app-hd4-parametrization}, both sides
are polynomials in $u$, and the identity is checked by exact expansion.
Combining it with the SOS certificate gives
\begin{equation}
 16\operatorname{per}(A)-\sum_{j=1}^{4}S_j(A)
 =\frac{2}{3}Q_4(A)\geq0.
 \label{eq:app-hd4-average}
\end{equation}
Choose a column $j_*$ minimizing $S_j(A)$. Then
\begin{equation}
 S_{j_*}(A)\leq\frac14\sum_{j=1}^{4}S_j(A)
 \leq4\operatorname{per}(A),
 \label{eq:app-hd4-merris-conclusion}
\end{equation}
which proves Eq.~\ref{eq:app-merris-four-result}. This argument applies
directly to every $A\in\Omega_4$ and does not require a limiting argument.

\paragraph{Formal verification and the public theorem's scope.}
The supplied proof record describes the \texttt{HolensDjokovic44Fast}
project using Lean and Mathlib version \texttt{v4.30.0-rc1}.
Its sparse-polynomial modules establish evaluation semantics, while the
symmetry and multiplier lemmas justify nonnegativity of each orbit
certificate. Each slice checks rational weight nonnegativity and an
exact polynomial identity. The certificate module combines all four
blocks and identifies their sum with
$24\operatorname{per}(A(u))-2-\sum_{i,j}a_{ij}(u)^3$.
The final matrix theorem connects the parametrization to Mathlib's
\texttt{Matrix.permanent} for arbitrary $A\in\Omega_4$.

The public theorem \texttt{holens\_djokovic\_4\_4} states
Eq.~\ref{eq:app-hd4-target}. The public theorem \texttt{merris\_n4}
assumes, in addition to $A\in\Omega_4$, that
\[
 S_4(A)\leq S_1(A),\qquad S_4(A)\leq S_2(A),\qquad S_4(A)\leq S_3(A),
\]
and concludes $S_4(A)\leq4\operatorname{per}(A)$.
For the mathematical existential statement, any minimizing column can
be moved to position four by a column permutation. Such a permutation
preserves double stochasticity and the permanent, and permutes the
values $S_j$. Applying the stated theorem to the permuted matrix and
relabeling the column yields Eq.~\ref{eq:app-merris-four-result}.
The supplied record documents the fixed-minimum-column Lean theorem;
it does not list a separate public theorem with the existential conclusion.

The record reports no \texttt{sorry}, \texttt{admit}, or project-defined
axioms. Large finite equalities are checked with \texttt{native\_decide},
whose trusted base includes Lean's native code generator and runtime.

\end{document}